\documentclass{article}

\usepackage[final]{corl_2026} % Uncomment for the camera-ready ``final'' version.
\usepackage{booktabs}
\usepackage{graphicx}
\usepackage{caption}
\usepackage{amsmath}
\usepackage{amssymb}
\usepackage{algorithm}
\usepackage{algpseudocode}
\usepackage{makecell}
\usepackage{enumitem}
\usepackage{amsthm}
\usepackage{mathtools}
\usepackage{needspace}
\usepackage{wrapfig}
\usepackage{dsfont}
\usepackage{xcolor}

\newtheorem{thmA}{Theorem}[section]
\newtheorem{propA}[thmA]{Proposition}
\newtheorem{lemA}[thmA]{Lemma}
\newtheorem{corA}[thmA]{Corollary}
\newtheorem{defnA}[thmA]{Definition}
\newtheorem{assA}{Assumption}
\newtheorem{remA}{Remark}

\newtheorem*{mainthm}{Theorem}

\newcommand{\R}{\mathbb{R}}
\newcommand{\norm}[1]{\left\|#1\right\|}
\newcommand{\op}{\mathrm{op}}
\newcommand{\DAP}{\mathrm{DAP}}
\newcommand{\FC}{\mathrm{FC}}
\newcommand{\expt}{\mathrm{exp}}
\newcommand{\refsym}{\mathrm{ref}}
\definecolor{stdgraycolor}{gray}{0.45}
\newcommand{\stdgray}[1]{{\tiny\textcolor{stdgraycolor}{$\pm #1$}}}

\newcommand{\Ext}{\mathcal{E}}
\newcommand{\Epiof}[1]{\Ext_\pi^{(#1)}}              % learned policy extrap at flow time #1
\newcommand{\Efof}[1]{\Ext_f^{(#1)}}                  % learned dynamics extrap at flow time #1
\newcommand{\Epistarof}[1]{\Ext_\pi^{*,(#1)}}        % ideal policy extrap at flow time #1
\newcommand{\Efstarof}[1]{\Ext_f^{*,(#1)}}            % ideal dynamics extrap at flow time #1
\newcommand{\Fof}[1]{F^{(#1)}}                        % learned alignment map at flow time #1
\newcommand{\Fstarof}[1]{F^{*,(#1)}}                  % ideal alignment map at flow time #1
\newcommand{\ostarof}[1]{o^{*,(#1)}}                  % fixed point of F^{(#1)}
\newcommand{\atauk}[1]{a^{(#1)}}                      % current action flow sample at time #1
\newcommand{\otauk}[1]{o^{(#1)}}                      % current future-obs flow sample at time #1
\newcommand{\ahattauk}[1]{\hat{a}^{(#1)}}             % extrapolated action estimate at step with time #1
\newcommand{\ohattauk}[1]{\hat{o}^{(#1)}}             % extrapolated future estimate at step with time #1
\newcommand{\uhata}[1]{\hat{u}^{(#1)}_a}              % learned policy field at flow time #1
\newcommand{\uhatf}[1]{\hat{u}^{(#1)}_o}              % learned dynamics field at flow time #1
\newcommand{\uideala}[1]{u^{*,(#1)}_a}                % ideal policy field at flow time #1
\newcommand{\uidealf}[1]{u^{*,(#1)}_o}                % ideal dynamics field at flow time #1

\hypersetup{pdftitle={Learning Robust Visuomotor Policies via Dynamics Alignment}, pdfauthor={Dohyeok Lee, Jung Min Lee, Munkyung Kim, Seokhun Ju, Jin Woo Koo, Kyungjae Lee, Dohyeong Kim, Taehyun Cho, Tapomayukh Bhattacharjee, Jungwoo Lee}}
\title{Learning Robust Visuomotor Policies via Dynamics Alignment}

\author{
    Dohyeok Lee$^1$, Jung Min Lee$^1$, Munkyung Kim$^1$, Seokhun Ju$^1$, Jin Woo Koo$^1$, Kyungjae Lee$^2$, \\
    \textbf{Dohyeong Kim}$^1$, \textbf{Taehyun Cho}$^1$, \textbf{Tapomayukh Bhattacharjee}$^3$, \textbf{Jungwoo Lee}$^1$ \\
  $^1$Seoul National University, $^2$Korea University, $^3$Cornell University\\
  \\
  \textbf{Project Page: \href{https://dynamics-aligned-policy.github.io}{\textcolor{magenta}{https://dynamics-aligned-policy.github.io}}}
}

\begin{document}
% \maketitle
% \thispagestyle{empty}
% \pagestyle{empty}

% \twocolumn[{
% \renewcommand\twocolumn[1][]{#1}   
% \maketitle
% \thispagestyle{empty}
% \pagestyle{empty}
% \vspace{-5mm}                       
% \begin{center}
%   \includegraphics[width=0.95\textwidth]{dynamics-aligned/icra/img/concept_v2_1.pdf}
%   \captionof{figure}{Comparison of three policy learning architectures for visuomotor control.
%   (a) \textbf{Diffusion Policy} iteratively denoise action sequences from pure noise $a_t^{(K)}$ to clean actions $a_t^{(0)}$ through $K$ reverse diffusion steps using observation conditioning without explicit dynamics modeling. 
%   (b) \textbf{Video-Conditioned Policy} leverages future observation predictions $\hat{o}_{t+1}$ or its latent $z(\hat{o}_{t+1})$ from a learned video prediction model $\epsilon_\phi$ to guide action generation using a diffusion model. 
%   However, it maintains separate video prediction and policy models during generation. 
%   (c) \textbf{Dynamics-Aligned Flow Matching Policy (ours)} integrates dynamics directly into the flow matching process of policy that progresses from noise to actions, learning a policy that maintains action consistency with predicted future observations.
%   }
%   \label{fig:concept}
% \end{center}
% }]
\makeatletter
% \g@addto@macro\@maketitle{
%     \begin{center}
%       \includegraphics[width=0.95\textwidth]{img/concept_v2_1.pdf}
%       \captionof{figure}{Comparison of three policy learning architectures for visuomotor control.
%   (a) \textbf{Diffusion Policy} iteratively denoise action sequences from pure noise $a_t^{(K)}$ to clean actions $a_t^{(0)}$ through $K$ reverse diffusion steps using observation conditioning without explicit dynamics modeling. 
%   (b) \textbf{Video-Conditioned Policy} leverages future observation predictions $\hat{o}_{t+1}$ or its latent $z(\hat{o}_{t+1})$ from a learned video prediction model $\epsilon_\phi$ to guide action generation using a diffusion model. 
%   However, it maintains separate video prediction and policy models during generation. 
%   (c) \textbf{Dynamics-Aligned Flow Matching Policy (ours)} integrates dynamics directly into the flow matching process of policy that progresses from noise to actions, learning a policy that maintains action consistency with predicted future observations.
% }
%       \label{fig:concept}
%     \end{center}
% }
\g@addto@macro\@maketitle{
    \begin{center}
      \includegraphics[width=0.95\textwidth]{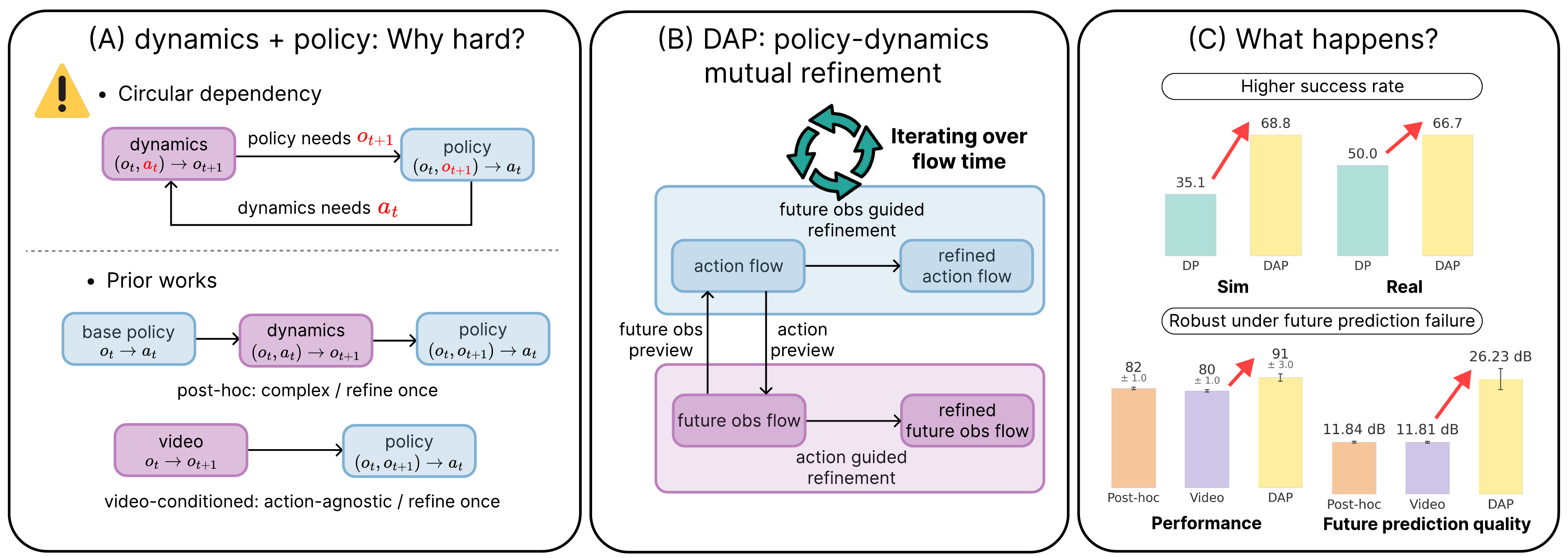}
      \captionof{figure}{Overview of Dynamics-Aligned Flow Matching Policy (DAP). Here, $o_t$, $a_t$, and $o_{t+1}$ denote observation, action, and future observation (future obs), respectively. 
      \textbf{(A)} Directly combining an action-conditioned dynamics model with a future-conditioned policy creates a circular dependency where dynamics needs actions to predict future obs, while the policy needs those future obs to choose actions.
      \textbf{(B)} DAP resolves this dependency by jointly sampling action and future obs flows, with extrapolated predictions from each flow iteratively refining the other.  
      \textbf{(C)} The final panel summarizes simulation, real-world, and future-prediction perturbation results.}
      \label{fig:concept}
    \end{center}
}
\makeatother

\maketitle
\thispagestyle{empty}
\pagestyle{empty}

% \begin{figure*}[!ht]
% % \vskip 0.2in
% \vspace{-10pt}
%     \centering
%     \includegraphics[width=\textwidth]{img/concept.pdf}
%     \caption{Overview of the proposed method.
%     The dynamics model predicts future observation from current observation and actions, while the policy model generates action sequences.
%     During generation, intermediate predictions by extrapolation from both models are used for mutual correction through iterative flow sharing.}
%     \label{fig:concept}
%     \vspace{-10pt}
% \end{figure*}

%===============================================================================

\begin{abstract}
    %The purpose of this document is to provide both the basic paper template and submission guidelines. Abstracts should be a single paragraph, between 4--6 sentences long, ideally. Gross violations will trigger corrections at the camera-ready phase.
    % Generative behavior-cloning policies such as diffusion and flow matching are effective for visuomotor manipulation, but they often fail to stay aligned with expert behavior when demonstration coverage is sparse.
    % To align action generation to expert behavior, we leverage an action-conditioned dynamics model that predicts the future observations a candidate action would induce.
    % However, this creates a circular dependency. 
    Generative BC policies model expert action distributions well, but they do not explicitly reason about what scene a candidate action would produce, so they cannot tell whether it reaches a task-relevant state. 
    We address this by coupling action generation with an action-conditioned dynamics model that predicts the future observations induced by actions. 
    However, naively placing such a dynamics model inside a generative policy creates a circular dependency.
    Future observations are needed to guide action generation, yet those observations can only be predicted from the action samples being generated.
    We introduce Dynamics-Aligned Flow Matching Policy (DAP), where a flow matching policy and an action-conditioned flow matching dynamics model co-sample actions and future observations through flow extrapolation.
    At each step the two flows correct each other, and the samples converge to a self-consistent action and future pair on which the policy and dynamics model agree.
    Across four simulated tasks and three real-world tasks, DAP achieves the highest average success rate, with relative improvements of 56.5\% in simulation and 33.3\% in the real world over the strongest baseline in each setting.
    Moreover, when the future-observation estimate is corrupted at inference time, DAP recovers a self-consistent action--future pair through iterative refinement, whereas video-conditioned and post-hoc dynamics baselines, which use a fixed future surrogate, degrade or fail.
\end{abstract}


%===============================================================================

\section{Introduction}
	
    %Submission to CoRL 2026 will be entirely electronic, via a web site (not email). Information about the submission process and \LaTeX{} templates are available on the conference web site at \url{https://corl.org/}. For camera ready submission, use the \texttt{final} option for the \texttt{\textbackslash usepackage} command. 

Behavior cloning (BC) has proven effective for learning robotic manipulation policies with diffusion-based generative models~\cite{DP,ditpolicy,scaledp}, yet these methods do not explicitly reason about the consequences of their actions.
% When expert trajectories sparsely cover the relevant state-action space, BC policies can overfit to demonstrated behaviors and generalize poorly beyond the training scenarios~\cite{vladiffusion,decomposing,stableBC,conBC}.
% A visuomotor policy should not only imitate expert actions; it should also know whether its candidate action will move the scene toward the next useful state.
% This action-consequence reasoning is especially important in limited-demonstration settings, where a single poor intermediate action can move the robot outside the support of expert data.
Future prediction therefore offers a useful inductive bias for visuomotor control.
A predicted observation helps judge whether candidate actions lead to task-relevant states.
Recent approaches~\cite{vidman, gr2, vpdd, susie, vpp, unipi, ivideogpt, seer} use video prediction or future representations as visual priors, often leveraging large-scale pretraining on datasets such as Open X-Embodiment (OXE)~\cite{oxe}.
However, action-agnostic methods model only the visual transition $p(o_{t+1}\mid o_t)$, so they only answer what typically follows rather than what a specific action would cause.
%and cannot answer what would happen if this specific action were executed.
%They answer what will happen next, not this action-specific counterfactual.
% Because expert videos typically cover successful trajectories, such models can inherit the same limited-support problem as BC and often rely on large-scale pretraining such as Open X-Embodiment (OXE)~\cite{oxe}.

Action-conditioned dynamics models~\cite{uva,hma,lpb} address this limitation more directly by learning action-conditioned visual dynamics $(o_t, a_t) \rightarrow o_{t+1}$ from the action labels already present in robot demonstrations.
Intuitively, conditioning on candidate actions turns future prediction from a passive forecast into a counterfactual query that asks what visual future this action would tend to induce given the current observation.
This action-consequence signal can complement imitation by providing an additional way to compare plausible actions beyond their likelihood under the expert dataset.

However, directly placing an action-conditioned dynamics model inside a generative policy is not straightforward.
The dynamics model needs candidate actions to predict future observations, while the policy should choose candidate actions by reasoning about those predicted observations, creating a \textbf{circular dependency}.
If dynamics is applied only after a full action sequence is sampled, it becomes a post-hoc evaluator rather than a mechanism that shapes generation~\cite{dreamtacvla,ladiwm,gpc,irasim}.
And if future observations are predicted before actions are available, the signal becomes action-agnostic again.
The central challenge is therefore to couple action generation with dynamics prediction while both action and future-observation samples are still evolving.

To address this challenge, we propose \textbf{Dynamics-Aligned Flow Matching Policy (DAP)}, which co-samples action trajectories and action-conditioned future observations.
Rather than aligning future latents through post-hoc guidance or an auxiliary training loss, DAP uses sampling-time co-generation for dynamics alignment.
Intermediate action samples pass through the dynamics model to imagine their visual consequences, and the resulting future-observation samples feed back to correct the policy flow.
As illustrated in Figure~\ref{fig:concept}, the coupled flows turn visual dynamics from a post-hoc evaluator into a sampling-time self-correction mechanism.

In simulation, on four ManiSkill3~\cite{maniskill3} tasks with 100 demonstrations per task, DAP improves average success from 43.9\% for the strongest baseline (fine-tuned $\pi_0$~\cite{pi0}) to 68.8\% and outperforms DP~\cite{DP}, LPB~\cite{lpb}, UVA~\cite{uva}, and $\pi_0$.
In the real world, across three manipulation tasks with 50--100 demonstrations per task, DAP improves average success from 50\% (DP) to 66.7\%.
Ablations show that the gain over BC comes from conditioning on the predicted future, and perturbation experiments show that co-sampling recovers when that future is corrupted, which fixed conditioning cannot.

Our contributions are: 
\begin{itemize}
\item \textbf{Dynamics-aligned flow generation.} We introduce an inference-time coupling mechanism that uses action-conditioned dynamics feedback to steer flow matching policy samples, rather than scoring completed trajectories post hoc.
\item \textbf{Analysis of iterative alignment.} Under a contraction-style assumption that holds empirically on all simulation tasks, we theoretically show that iterative dynamics feedback removes the mismatch left by one-shot future conditioning.
\item \textbf{Empirical validation of dynamics-aligned generation.} We evaluate DAP on four simulated and three real-world manipulation tasks, showing relative average success rate gains of 56.5\% in simulation and 33.3\% in the real world over the strongest baseline in each setting, with additional perturbation and demonstration-scaling analyses.
\end{itemize}

%===============================================================================

%\section{Citations}
%\label{sec:citations}

	%Citations can be made using either \textbackslash citep\{\} or \textbackslash citet\{\}, depending from the appropriateness. To avoid the citation moving to the next line, it is often a good practice to replace the space before with a tilde (\~{}) character.
	%Example 1: ``CoRL is the best conference ever~\citep{Gauss1857}.''
	%Example 2: ``\citet{Lagrange1788} proved, both theoretically and numerically, that CoRL is the best conference ever.''
	
%===============================================================================

\section{Related Work}
\label{sec:related_works}

Prior work related to DAP spans generative imitation policies, future-prediction priors for visuomotor control, and action-conditioned dynamics for model-guided generation.

\textbf{Generative Policies for Robot Imitation Learning.}
Diffusion-based policies~\cite{DP} model multi-modal action distributions for robot behavior cloning and have been extended with DiT backbones~\cite{ditpolicy,scaledp}.
Flow matching policies~\cite{flowmatching} provide another generative imitation learning approach, with applications to vision-language-action models, Riemannian manifolds, point clouds, and affordance-conditioned generation~\cite{pi0,rfmp,pointcloudflowmatching,affordanceflowmatching}.
These samplers are powerful, but they do not explicitly represent what future observation a sampled action would induce.
DAP adds this missing action-consequence feedback by using predicted future observations to correct action samples during flow generation.

\textbf{Future Prediction and Video Priors for Robot Policy Learning.}
Future prediction provides a natural visual prior because a good action should not only resemble expert controls, but also move the scene toward plausible and useful future observations.
Following advances in large-scale video generation~\cite{sora,moviegen}, recent robot-learning methods use pretrained video models, future prediction, future representations, or hidden-state alignment for policy learning~\cite{vidman,gr2,vpdd,susie,vpp,unipi,ivideogpt,seer,flare}.
These methods show the value of future information, but the signal is often action-agnostic or auxiliary.
DAP instead learns action-conditioned visual dynamics for inference-time self-correction.

\textbf{Action-Conditioned Dynamics and Model-Guided Policy Generation.}
Learned dynamics have long supported model predictive control~\cite{finn2017deep,madan2026tactic}, as well as planning and imagination in model-based reinforcement learning~\cite{dreamerv3,iris,diamond}, and recent methods incorporate dynamics through state-dynamics regularization, video world models, latent action models, or unified video-action modeling~\cite{uva,hma,dynamicsawarediffusion,statediffusion,gvp-wm,genie,lee2026mvplam}.
Other methods use predictive models to steer policies back toward expert behavior, refine initial action proposals, or score and guide candidate actions at inference time~\cite{lpb,dreamtacvla,ladiwm,gpc,irasim,dynaguide}.
Together, these approaches show that predictive models can improve policy generation beyond pure behavior cloning, but they typically operate as post-hoc refiners after an action proposal is produced or as an external guide.
DAP instead couples action and future-observation flows at every generative timestep, so intermediate actions induce future predictions that immediately correct the policy flow and resolve the circular dependency between action generation and dynamics prediction.

\section{Method}
\label{sec:method}
% Our approach consists of two key innovations: (1) explicit dynamics modeling using flow matching, and (2) iterative coupling between dynamics and policy generation. Figure~\ref{fig:arch} presents our architectural overview, and we detail the design in Section~\ref{sec:method}.

\subsection{Preliminaries} \label{sec:preliminary}
\subsubsection{Problem Formulation}
We consider behavior cloning from expert demonstrations $\mathcal{D}_{\text{expert}}$. The policy models $\pi(a_t|o_t)$, and the action-conditioned dynamics model predicts $p(o_{t+1}|o_t,a_t)$, where $o_t$, $a_t$, and $o_{t+1}$ denote chunked observation, action, and future-observation variables.
Fixed conditioning (FC) denotes methods that compute one future surrogate $\hat{o}^{\FC}$ before action sampling and keep it fixed.
DAP instead updates the action and future estimates jointly at every flow step.
A hat denotes a model-based prediction, including an extrapolated endpoint estimate.
For discrete sampling, we use $K$ flow steps with step size $d\tau=1/K$.

\begin{figure*}[!t]
    \centering
    \includegraphics[width=0.95\textwidth]{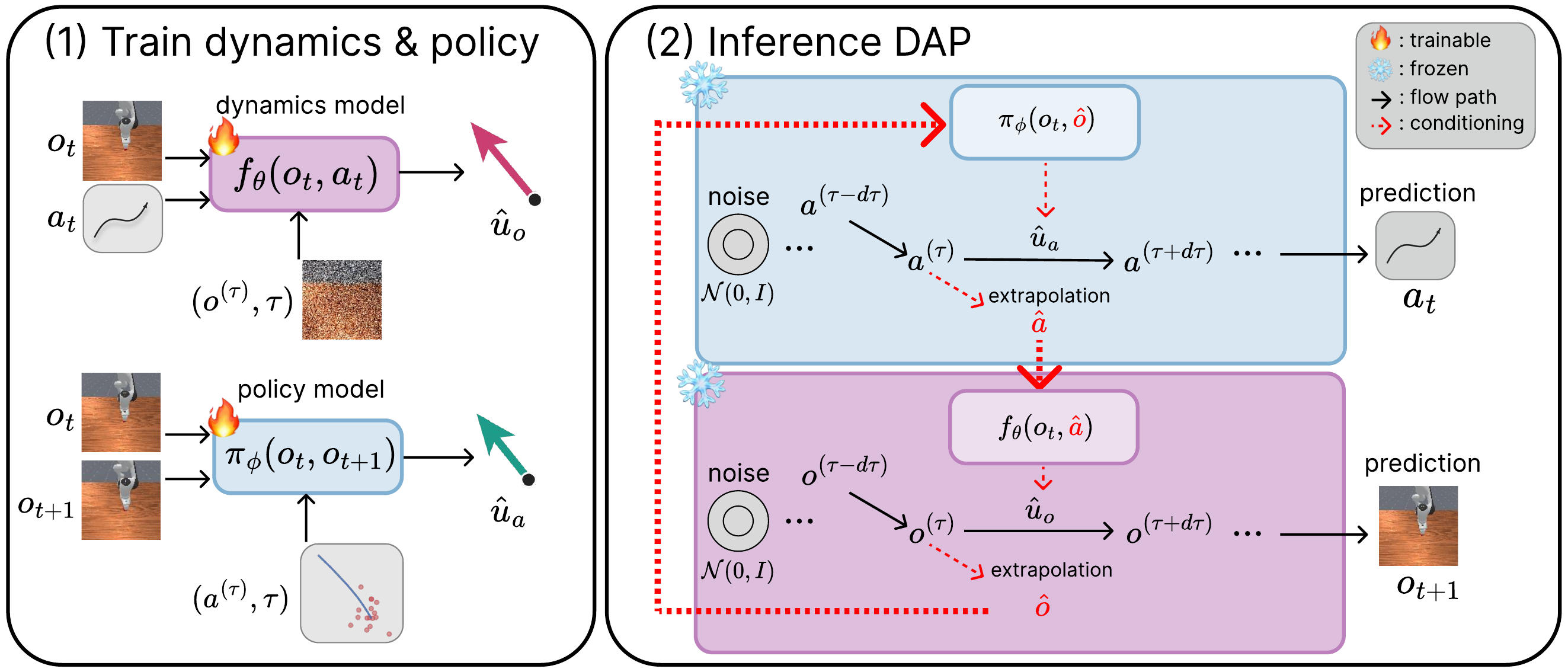}
    \caption{Training and inference pipeline for DAP. \textbf{(1)} During training, DAP learns an action-conditioned dynamics flow and a future-conditioned policy flow. \textbf{(2)} During inference, the two frozen flows exchange extrapolated action and future-observation estimates, iteratively co-refining mutually aligned samples.}
    \label{fig:dap_detail}
\end{figure*}

% Our training dataset consists of two types of trajectories: expert demonstrations $\mathcal{D}_{\text{expert}} = \{(o_t, a_t, o_{t+1})\}$ with $N_e$ episodes collected from human operators, and random exploration data $\mathcal{D}_{\text{random}} = \{(o_t, a_t, o_{t+1})\}$ with $N_r$ episodes collected via random policies.
% While expert data provides high-quality observation-action pairs for policy learning, random data expands the observation-action coverage to improve generalization, particularly for dynamics modeling.
% The combined dataset $\mathcal{D}_{\text{expert}} \cup \mathcal{D}_{\text{random}}$ enables our approach to leverage both demonstration quality and data diversity.

\subsubsection{Flow Matching Formulation}
We use flow matching~\cite{flowmatching} to train both the policy and dynamics model.
Detailed background is provided in Appendix~\ref{app:flow_matching_formulation}.
Let $x^{(\tau)}$ denote a flow sample at flow timestep $\tau \in [0,1]$, where the path transforms Gaussian noise ($\tau=0$) into data ($\tau=1$).
We use $a^{(\tau)}$ for the action flow sample and $o^{(\tau)}$ for the future-observation flow sample, with the trajectory subscripts $t$ and $t+1$ omitted for brevity.
With the standard linear interpolation between a data sample $x_1$ and noise $x^{(0)} \sim \mathcal{N}(0,I)$,
\begin{equation}
    x^{(\tau)} = \tau x_1 + (1-\tau)x^{(0)}, \quad x^{(0)} \sim \mathcal{N}(0, I)
\end{equation}
the conditional vector field is $u^{(\tau)}(x^{(\tau)} | x_1)=x_1-x^{(0)}$.
For this linear path, the target value is constant along each sampled trajectory.
The superscript $(\tau)$ indicates that the vector field is evaluated at flow time $\tau$ through the sample $x^{(\tau)}$.
Conditional flow matching trains a neural vector field to regress this target:
\begin{equation}
    \mathcal{L}_{\text{CFM}}(\theta) = \mathbb{E}\left[\lVert u^{(\tau)}(x^{(\tau)} | x_1) - v_\theta(x^{(\tau)}, \tau)\rVert^2\right]
    \label{eq:cfm_objective}
\end{equation}
In the main text, $\pi_\phi$ and $f_\theta$ denote the neural networks, and their realized velocity outputs are written as $\hat{u}_a^{(\tau)}$ and $\hat{u}_o^{(\tau)}$.

\textbf{Flow Extrapolation.}
To provide more reliable conditioning samples to both models, we \textit{extrapolate} the current flow sample $x^{(\tau)}$ using the model's velocity estimate $\hat{u}^{(\tau)}$, following the approach proposed in Rectified Flow~\cite{rectflow}. 
\begin{equation}
    \mathcal{E}(x^{(\tau)}) \coloneqq x^{(\tau)} + (1-\tau)\,\hat{u}^{(\tau)}(x^{(\tau)})
\end{equation}
For the policy flow, this gives $\mathcal{E}(a^{(\tau)})$ using the output of $\pi_\phi$.
For the dynamics flow, it gives $\mathcal{E}(o^{(\tau)})$ using the output of $f_\theta$.
In the inference stage, $\mathcal{E}(x^{(\tau)}) \approx x_1$ and this approximation becomes accurate if the learned vector field predicts $u^{(\tau)}(x^{(\tau)} | x_1)$ accurately. 
%By extrapolating to the end of the flow trajectory, we obtain a more stable sample to condition the other model compared to using the noisy flow sample $x^{(\tau)}$. 
Extrapolating each intermediate flow sample to the trajectory endpoint provides more accurate corrective feedback between the policy and dynamics models.
See Appendix~\ref{app:flow_extrapolation_eval} for the flow extrapolation experiments and results.

\subsection{Dynamics-Aligned Flow Matching Policy}\label{sec:dap}
In this paper, dynamics alignment refers specifically to generation-time coupling between the policy flow and the explicit dynamics flow, not to latent or representation alignment imposed by an auxiliary loss.
DAP uses two conditional flow models that are trained separately but coupled at inference time (Figure~\ref{fig:dap_detail}).
The dynamics model predicts action-conditioned future observations, while the policy model generates actions conditioned on current and predicted future observations.
The goal is to sample a self-consistent action--future-observation pair rather than using a fixed future surrogate produced outside the policy sampling loop.

\paragraph{Training Objectives.}
%Let $f_\theta$ denote the action-conditioned dynamics model and $\pi_\phi$ denote the future-conditioned policy model.
% In the dynamics model training, we leverage both expert dataset $\mathcal{D}_\text{expert}$ and the random dataset $\mathcal{D}_{\text{random}}$.
% This is because the $\mathcal{D}_\text{expert}$ is a collection of successful demonstrations of tasks which includes only a small subset of the action distribution.
% We collect the random dataset $\mathcal{D}_\text{random}$ with the same size as the expert demonstrations $\mathcal{D}_\text{expert}$, to sufficiently support the broad observation-action space. 
% This method allows the dynamics model to provide the future observation with high fidelity, even when the action predicted by the policy substantially deviates from the expert distribution.
%hen, we train the dynamics model $f_\theta$ on a combined dataset $\mathcal{D}_\text{total}$ using the following objective:
We first train the action-conditioned dynamics model $f_\theta$ using the following flow matching objective.
Specifically, for $\epsilon_o \sim \mathcal{N}(0, I)$, $\tau \sim p(\tau)$, and $(o_t, a_t, o_{t+1}) \sim \mathcal{D}_{\text{expert}}$, 
\begin{equation}
\mathcal{L}_{\text{dyn}}(\theta) =\|u_o - f_\theta(o^{(\tau)}, \tau | o_t, a_t)\|^2
\label{eqn:dyn_obj}
\end{equation}
%where $z_t$ and $z_{t+1}$ are the latent embeddings of $o_t$ and $o_{t+1}$, respectively.
The flow matching targets are defined as $u_o = o_{t+1} - \epsilon_o$ and $o^{(\tau)} = \tau o_{t+1} + (1-\tau) \epsilon_o$.
We then train policy $\pi_{\phi}$ on the expert dataset using an analogous flow matching objective.
For $\epsilon_a \sim \mathcal{N}(0, I)$, $\tau \sim p(\tau)$, and $(o_t, a_t, o_{t+1}) \sim \mathcal{D}_{\text{expert}}$,
\begin{equation}
\mathcal{L}_\pi(\phi) = \|u_a - \pi_\phi(a^{(\tau)}, \tau | o_t, o_{t+1})\|^2
\label{eqn:pi_obj}
\end{equation}
where $u_a = a_t - \epsilon_a$ and $a^{(\tau)} = \tau a_t + (1-\tau) \epsilon_a$.
%We provide the pseudo-code in Algorithm \ref{alg:train}.

% \paragraph{Training Details}
% We provide the implementation details of the key models: the dynamics model $f_\theta$ and the policy model $\pi_\phi$.

\textbf{Implementation.} The dynamics model uses a DiT architecture~\cite{dit} with SD-VAE~\cite{SD1} latents and action cross-attention.
For the simulation tasks, we fine-tuned the SD-VAE to reduce the domain gap.
The policy follows the Diffusion Policy U-Net architecture with cross-attention layers for future-observation conditioning, and is trained on ground-truth futures with 0.1 future-conditioning dropout.
Full architecture and preprocessing details are in Appendix~\ref{app:implementation_details}, and Appendix~\ref{app:train_test_shift} shows that conditioning on predicted rather than ground-truth futures at test time barely shifts the executed action.

\paragraph{Dynamics-Aligned Sampling.}
Dynamics-policy coupling is needed because the two models are mutually dependent.
The dynamics model $f_\theta$ learns $(o_t, a_t) \rightarrow o_{t+1}$ and requires a candidate action, whereas the policy model $\pi_\phi$ benefits from conditioning on the future consequences of its candidate actions.
In a sequential procedure, where the policy first generates actions and only then feeds them to the dynamics model, the policy cannot use this feedback because action generation has already finished.
DAP initializes both an action flow $a^{(0)}$ and a future-observation flow $o^{(0)}$ from Gaussian noise, and evolves them in parallel over the discretized flow schedule $\tau \in \{0,d\tau,2d\tau,\ldots,1-d\tau\}$ with step size $d\tau=1/K$.
Alongside the samples, DAP keeps endpoint estimates $\hat{a}$ and $\hat{o}$, which are overwritten at every step and initialized by extrapolating $a^{(0)}$ and $o^{(0)}$ without cross-conditioning.
One coupled update at flow time $\tau$ is
\begin{align}
    \hat{u}_a^{(\tau)}
    &= \pi_\phi(a^{(\tau)}, \tau \mid o_t, \hat{o}),
    &
    \hat{u}_o^{(\tau)}
    &= f_\theta(o^{(\tau)}, \tau \mid o_t, \hat{a}),
    \nonumber \\
    a^{(\tau+d\tau)}
    &= a^{(\tau)} + d\tau\,\hat{u}_a^{(\tau)},
    &
    o^{(\tau+d\tau)}
    &= o^{(\tau)} + d\tau\,\hat{u}_o^{(\tau)},
    \nonumber \\
    \hat{a}
    &\leftarrow a^{(\tau)} + (1-\tau)\hat{u}_a^{(\tau)},
    &
    \hat{o}
    &\leftarrow o^{(\tau)} + (1-\tau)\hat{u}_o^{(\tau)}.
    \label{eq:dap_sampling_step}
\end{align}
The last line is the flow-extrapolation step. 
It maps noisy intermediate samples to endpoint estimates that can condition the other model at the next step.
Repeating this exchange turns the dynamics model from a post-hoc evaluator into a sampling-time feedback signal, so intermediate action intentions update the dynamics prediction and the updated future prediction refines the policy flow.
The pseudo-code is provided in Algorithm~\ref{alg:sample}.

We also provide a theoretical analysis showing how dynamics-aligned sampling removes the fixed-conditioning residual under the stated contraction assumptions.
The core mechanism is that applying the dynamics-alignment correction at each flow step defines a contraction on the conditioning mismatch, progressively reducing the gap between the assumed and actual future.
A detailed derivation and full proofs for the sampler of Algorithm~\ref{alg:sample} are provided in Appendix~\ref{app:formal-theorem}, which also shows that the contraction assumption holds empirically on all simulation tasks.

\needspace{13\baselineskip}
\begin{mainthm}[Dynamics-Aligned Contraction; informal]
\label{thm:main-informal}
Under regularity and contraction conditions on the learned models
$($Assumptions~\ref{ass:cross-lip}--\ref{ass:contraction},
Appendix~\ref{app:formal-theorem}$)$,
let $a^{\expt}$ denote the expert action and $o^{\expt}$
the expert future observation.
Then the following bounds hold:
\begin{align}
    \textbf{DAP:}\quad
    \norm{\hat{o}^{\DAP} - o^{\expt}}
    &\leq \underset{\to\,0}{\underbrace{\mathcal{O}(1/K)}}\,,
    &
    \norm{\hat{a}^{\DAP} - a^{\expt}}
    &\leq \underset{\to\,0}{\underbrace{\mathcal{O}(1/K)}}
    + \mathcal{O}(\varepsilon),
    \label{eq:main-dap} \\[6pt]
    \textbf{FC:}\quad
    \norm{\hat{o}^{\FC} - o^{\expt}}
    &= \underset{>\,0}{\underbrace{\mathrm{const.}}}\,,
    &
    \norm{\hat{a}^{\FC} - a^{\expt}}
    &\leq \underset{\not\to\,0}{\underbrace{C_{\FC}}}
    + \mathcal{O}(\varepsilon),
    \label{eq:main-fc}
\end{align}
where $K=1/d\tau$ is the number of flow steps, $\varepsilon$ is the model approximation error, and $C_{\FC} > 0$ is the non-vanishing conditioning mismatch of fixed conditioning, which keeps a single future-observation surrogate throughout generation.
\end{mainthm}

Thus, the theorem shows that DAP benefits from more flow steps by contracting the action--future mismatch and eliminating the tracking residual at rate $\mathcal{O}(1/K)$ up to the model-error floor $\mathcal{O}(\varepsilon)$, whereas FC retains a fixed residual because its future prediction is not updated by action feedback.

\section{Experimental Results}
\label{sec:result}

\begin{figure}[!t]
    \centering
    \includegraphics[width=\textwidth]{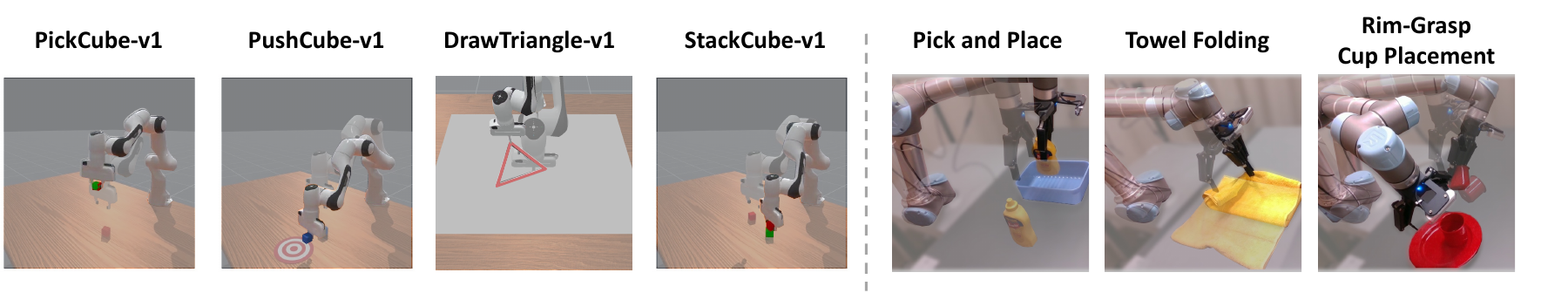}
    \caption{Benchmark simulation and real-world manipulation tasks used in our experiments.}
    \label{fig:task_vis}
\end{figure}

Our evaluation focuses on two questions: (i) Does DAP improve policy performance, and which component drives the gain? (ii) Does iterative co-sampling recover when future-observation estimates are corrupted at inference time?
We benchmark DAP on diverse simulation and real-world manipulation tasks (Figure~\ref{fig:task_vis}) and compare it against strong baselines.
\paragraph{Baselines.}
For fair comparison, we compare against DP~\cite{DP}, LPB~\cite{lpb}, UVA~\cite{uva}, and fine-tuned $\pi_0$~\cite{pi0} using official implementations under matched expert demonstration budgets.
DP is a diffusion-based BC policy, LPB applies dynamics-barrier guidance to diffusion-policy rollouts, UVA is a unified video-action model, and $\pi_0$ is a pretrained VLA model fine-tuned with LoRA~\cite{lora}. 
We use the settings from each official implementation as provided, and keep all remaining configurations identical across methods.
Whenever a baseline includes separate dynamics/video and policy stages, those stages use the same epoch budgets as the corresponding DAP components.
Checkpointing, hyperparameters, and implementation deviations are reported with each method in Appendix~\ref{app:implementation_details}.

\subsection{Simulation Experiments}
\paragraph{Experimental Procedure.}
For evaluation, we benchmark DAP and baselines on four manipulation tasks in the ManiSkill3~\cite{maniskill3} environment. 
For each task, we only use 100 expert demonstrations to train all methods and evaluate each method over 100 trials.

\paragraph{Results.}
\begin{table}[h]
    \centering
    %\footnotesize
    \setlength{\tabcolsep}{3pt}
    \begin{tabular}{@{}c|cccc|c@{}} \toprule
    Tasks & PickCube-v1 & PushCube-v1 & DrawTriangle-v1 & StackCube-v1 & Average \\ \midrule
    \# of Demos   & 100    & 100         & 100             & 100            & -- \\ \midrule
    DP      & 78/52.0\,\stdgray{12.3} & 100/82.3\,\stdgray{9.3} & 12/6.0\,\stdgray{5.6} & 2/0.0\,\stdgray{0.0} & 48/35.1\,\stdgray{2.7} \\
    LPB     & 47/46.0\,\stdgray{0.0} & 100/\textbf{100.0}\,\stdgray{0.0} & 22/18.7\,\stdgray{3.1} & 12/8.3\,\stdgray{1.2} & 45/43.3\,\stdgray{0.5} \\
    UVA     & 7/4.3\,\stdgray{0.6} & 78/75.3\,\stdgray{0.6} & 45/44.0\,\stdgray{0.0} & 3/0.7\,\stdgray{0.6} & 33/31.1\,\stdgray{0.1} \\
    $\pi_0$ & 42/34.0\,\stdgray{1.0} & 100/98.3\,\stdgray{0.6} & 10/2.0\,\stdgray{1.0} & \textbf{56}/\textbf{41.3}\,\stdgray{3.1} & 52/43.9\,\stdgray{0.9} \\
    DAP     & \textbf{94}/\textbf{88.7}\,\stdgray{0.6} & 100/98.3\,\stdgray{0.6} & \textbf{57}/\textbf{55.7}\,\stdgray{2.3} & 40/32.3\,\stdgray{4.0} & \textbf{73}/\textbf{68.8}\,\stdgray{1.6} \\ \bottomrule
    \end{tabular}
    \vspace{+5pt}
    \caption{Success rates (\%) on ManiSkill3 under the 100-demonstration setting, reported as best checkpoint / mean over the last three checkpoints $\pm$ standard deviation (gray). Best is \textbf{bolded}.}
    \label{tab:sim_main}
\end{table}
\vspace{-10pt}

% \begin{table}[h]
%     \centering
%     \begin{tabular}{@{}c|ccccc@{}} \toprule
%                     & DP    & LPB  & UVA   & $\pi_0$  & DAP \\ \midrule
%     Demos           & 100   & 100  & 100   & 100      & 100    \\ \midrule
%     PickCube-v1     & 0.78  & 0.47 & 0.07  & 0.42     & \textbf{0.96}    \\
%     PushCube-v1     & 1.00  & 1.00 & 0.78  & 1.00     & \textbf{1.00}    \\
%     DrawTriangle-v1 & 0.12  & 0.00 & 0.45  & 0.10     & \textbf{0.57}    \\
%     StackCube-v1    & 0.02  & 0.12 & -     & -        & \textbf{0.32} \\ \midrule
%     Average         & 0.48  & 0.40 & -     & -        & \textbf{0.71}    \\ \bottomrule
%     \end{tabular}
%     \vspace{+5pt}
%     \caption{Success rates on ManiSkill3 under the 100-demonstration setting, averaged over 100 evaluation trials. For DP DrawTriangle-v1, baseline numbers are reproduced in our setup when official results are unavailable.}
% \end{table}

DAP achieves the highest average success rate (Table~\ref{tab:sim_main}). Using the mean over the last three checkpoints, DAP improves over the strongest baseline $\pi_0$ from 43.9\% to 68.8\%, a 56.5\% relative improvement.
The gains are especially clear on PickCube-v1 and DrawTriangle-v1, where DAP produces action samples whose predicted visual consequences remain consistent with the intended task outcome.
DAP trails $\pi_0$ on StackCube-v1, and PushCube-v1 is saturated.
At inference time, DAP achieves lower action-generation latency than DP.
Detailed measurements are provided in Appendix~\ref{app:inference_time}.
%From a deployment perspective, DAP also remains practical in the simulated control loop: although dynamics alignment adds one dynamics-model forward pass alongside the policy forward pass at each flow step, the efficient sampling enabled by flow matching yields lower end-to-end action-generation latency than DP; detailed measurements are provided in Appendix~\ref{app:inference_time}.

\begin{figure}[!t]
    \centering
    \includegraphics[width=0.95\textwidth]{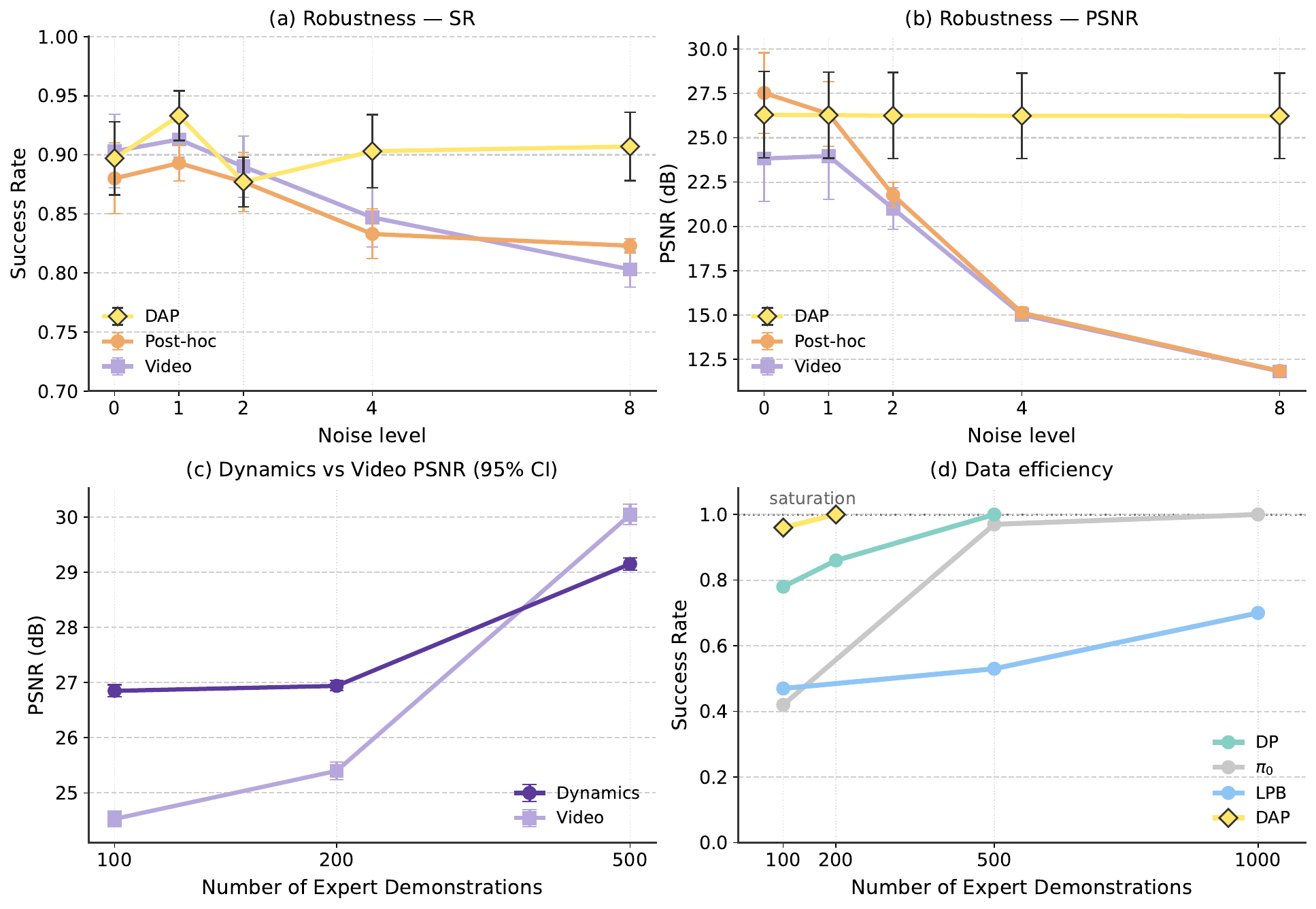}
    \caption{Robustness, prediction-quality, and data-efficiency analyses on PickCube-v1. Error bars denote checkpoint standard deviations in (a,b) and validation-set confidence intervals in (c). No error bars in (d).}
    \label{fig:ablation}
\end{figure}

\begin{wraptable}[8]{r}{0.43\textwidth}
    \vspace{-4pt}
    \centering
    \small
    \setlength{\tabcolsep}{3pt}
    \begin{tabular}{@{}l|cc@{}} \toprule
    Tasks           & DAP & DAP (w/o Dyn.) \\ \midrule
    PickCube-v1     & \textbf{88.7}\,\stdgray{0.6} & 73.7\,\stdgray{1.5} \\
    DrawTriangle-v1 & \textbf{55.7}\,\stdgray{2.3} & 22.3\,\stdgray{1.2} \\
    StackCube-v1    & \textbf{32.3}\,\stdgray{4.0} & 15.0\,\stdgray{1.0} \\ \bottomrule
    \end{tabular}
    \vspace{-2pt}
    \caption{Simulation dynamics ablation. Success rates (\%), mean over the last three checkpoints $\pm$ standard deviation.}
    \label{tab:sim_dyn_ablation}
    \vspace{-10pt}
\end{wraptable}

\paragraph{Dynamics Ablation.}
Table~\ref{tab:sim_dyn_ablation} compares DAP with DAP (w/o Dyn.), which uses the same flow matching policy without the dynamics model, on the non-saturated simulation tasks.
Removing the dynamics model lowers success on every task, with relative drops of 16.9\% to 59.9\%, so the gain does not come from the flow matching policy alone.
PushCube-v1 is omitted since DAP is already saturated on it.

% \paragraph{Does Dynamics Model Really Understand the Actions?}
% Following the $\Delta_\text{PSNR}$ protocol of Genie~\cite{genie}, we further test whether the learned dynamics model is action-sensitive rather than merely visually plausible. On PickCube-v1, we compare future-observation PSNR under the ground-truth action with the average PSNR under randomly sampled actions, using their gap as $\Delta_\text{PSNR}$. DAP yields a large gap of $7.19 \pm 1.79$ and maintains strong rollout PSNR under diffusion-policy actions, indicating that its predictions are meaningfully conditioned on control inputs and remain reliable under policy-induced behavior. Detailed protocol and full results are provided in Appendix~\ref{app:delta_psnr}.

\paragraph{Robustness to Corrupted Future Predictions.}
Figure~\ref{fig:ablation}(a) and \ref{fig:ablation}(b) evaluate whether DAP can recover when the future-observation conditioning used by each method is corrupted at inference time.
We corrupt the future-observation signal consumed by each method at its point of use, using Gaussian perturbations with the same relative noise level.
For DAP, this signal is the initial future-observation estimate $\hat{o}$, whereas for fixed-conditioning baselines it is the precomputed surrogate $\hat{o}^{\FC}$.
Because DAP's first policy update already conditions on the corrupted $\hat{o}$, the corruption enters both the action and future-observation flows.
We compare DAP with two fixed-conditioning baselines.
\textit{Post-hoc} first samples an action using DP, predicts $\hat{o}^{\FC}$ with the action-conditioned dynamics model, and then keeps it frozen during policy sampling.
\textit{Video} conditions the policy on an action-agnostic video prediction.
DAP maintains a success rate of 88--93\% and keeps the final prediction PSNR nearly constant around 26 dB across all noise levels, whereas both fixed-conditioning baselines suffer monotonic PSNR collapse and lower success at high noise. 
The stable final PSNR thus indicates recovery during generation, as iterative mutual correction pulls the sample back toward a self-consistent action--future pair.
At zero noise the fixed-conditioning baselines match DAP in success rate, so the benefit of co-sampling appears when the future estimate is unreliable.
At noise level 8, a frozen variant that shares DAP's weights but never re-estimates the future stays at 11.6~dB PSNR, while DAP recovers from 11.5~dB to its uncorrupted level of about 26~dB, so the recovery comes from in-loop re-estimation.

\begin{wrapfigure}{r}{0.5\textwidth}
    \vspace{-4pt}
    \centering
    \includegraphics[width=\linewidth]{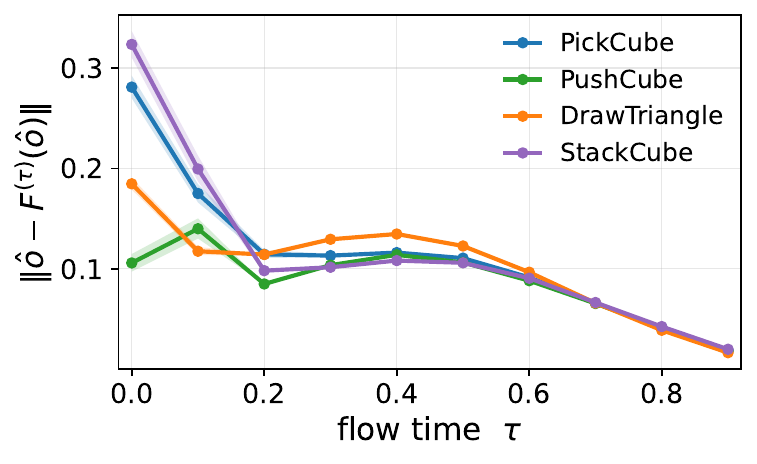}
    \vspace{-16pt}
    \caption{Alignment error $R^{(\tau)}(\hat{o})$ of DAP over flow time on the four simulation tasks, with 95\% bootstrap confidence intervals over 200 held-out states.}
    \label{fig:alignment_error}
    \vspace{-10pt}
\end{wrapfigure}

\paragraph{Alignment during Sampling.}
Figure~\ref{fig:alignment_error} plots the alignment error $R^{(\tau)}(\hat{o})=\norm{\hat{o}-F^{(\tau)}(\hat{o})}$, where $F^{(\tau)}(\hat{o})=\Efof{\tau}\bigl(\Epiof{\tau}(\hat{o})\bigr)$ extrapolates the action given $\hat{o}$ with the policy and then the future given that action with the dynamics model:
\begin{align*}
    \Epiof{\tau}(\hat{o}) &= a^{(\tau)} + (1-\tau)\,\pi_\phi(a^{(\tau)}, \tau \mid o_t, \hat{o}),\\
    \Efof{\tau}(\hat{a}) &= o^{(\tau)} + (1-\tau)\,f_\theta(o^{(\tau)}, \tau \mid o_t, \hat{a}).
\end{align*}
Within the first two flow steps, the error reaches a common level on all four tasks regardless of its initial value, which the $(1-\tau)$ factor in Eq.~\eqref{eq:dap_sampling_step} alone cannot explain, and it decreases to about 0.02 by the last step.
The measured contraction constants are reported in Appendix~\ref{app:measured}.

\paragraph{Scaling with Expert Demonstrations.}
Figure~\ref{fig:ablation}(c) and \ref{fig:ablation}(d) examine how future-prediction quality and task success scale with the number of expert demonstrations. 
First, for future-observation prediction, we train the same flow matching predictor either as an action-conditioned dynamics model or as an action-agnostic video model on the same PickCube-v1 demonstration datasets. 
The video predictor shows lower future-prediction quality at 100 and 200 demonstrations, trailing dynamics by 2.32 dB and 1.54 dB, respectively, and surpasses the dynamics model at 500 demonstrations.
PSNR alone, however, does not reflect action sensitivity: at every scale, the dynamics model's PSNR under the true action exceeds that under random actions by at least 4.75~dB, compared with near zero for the video model (Table~\ref{tab:psnr}), so the future supplied during co-sampling tracks the action being generated.
The evaluation protocol is detailed in Appendix~\ref{app:delta_psnr}.

\begin{wraptable}{r}{0.49\textwidth}
    \vspace{-8pt}
    \centering
    \small
    \setlength{\tabcolsep}{3pt}
    \begin{tabular}{@{}l|ccc@{}} \toprule
    \# of Demos & 100 & 200 & 500 \\ \midrule
    Dynamics    & $+4.75$\,\stdgray{0.09} & $+5.09$\,\stdgray{0.09} & $+6.29$\,\stdgray{0.10} \\
    Video       & $-0.06$\,\stdgray{0.10} & $+0.01$\,\stdgray{0.10} & $-0.00$\,\stdgray{0.10} \\ \bottomrule
    \end{tabular}
    \vspace{-2pt}
    \caption{$\Delta_\text{PSNR}$ (dB) on PickCube-v1, PSNR under the true action minus mean PSNR under 10 random actions, $\pm$ 95\% bootstrap confidence interval over 1600 test samples.}
    \label{tab:psnr}
    \vspace{-10pt}
\end{wraptable}

Second, using the best checkpoint, DAP reaches a near-saturated success rate with 100 demonstrations and saturates by 200 demonstrations, whereas the baselines improve more gradually as additional demonstrations are added. 
These trends suggest that, in this benchmark, action conditioning provides a stronger future-prediction signal at 100--200 demonstrations, and DAP translates this signal into faster task-success saturation.

% \begin{figure}[!t]
%     \centering
%     \includegraphics[width=\textwidth]{img/ablation_v2_3.pdf}
%     \caption{TBD}
%     \label{fig:ablation}
% \end{figure}

\subsection{Real-World Experiments}
% \begin{figure*}[t]
% % \vskip 0.2in
% % \vspace{-10pt}
%     \centering
% %     % \includegraphics[width=\textwidth]{img/main_res.pdf}
%     \includegraphics[width=\textwidth]{img/real_world_exp_v2.pdf}
% %     %\vspace{-10pt}
%     \caption{\textbf{Top:} Success rates across four real-world manipulation tasks. 
%     DAP achieves an average success rate of 75.0\%, outperforming DP (62.5\%), UVA (32.5\%), and $\pi_0$ (30.0\%). 
%     Notably, DAP is the only method that successfully performs the challenging Rim-Grasp Cup Placement task (30\%). 
%     \textbf{Bottom:} task setup for each manipulation scenario. \textbf{Lower Right}: trainable parameters comparison showing DAP (369M) maintains comparable model size to baselines.
%     }
%     \label{fig:main_result}
%     \vspace{-10pt}
% \end{figure*}

\paragraph{Experimental Procedure.}
%We evaluate our method using both success rate and action prediction error metric on the validation dataset. 
For real-world experiments, we use a UR5 robot equipped with two front-view RGB cameras and one wrist-mounted egocentric RGB camera.
Each method is tested across 10 trials per task. 
Demonstration counts denote teleoperated expert trajectories. 
Every policy including DAP is trained only on the shared expert demonstrations, so the policy comparison stays matched. 
Additional random rollouts are used only for the dynamics model, where they are cheap to collect without teleoperation and widen the action distribution it sees (Appendix~\ref{app:implementation_details}).
We evaluate Pick and Place (PP), Towel Folding (TF), and Rim-Grasp Cup Placement (RCP) as three real-world manipulation tasks. 
RCP requires grasping a horizontally placed cup by its rim and placing it upright on a plate, where success is defined by stable placement in the target region. 
Compared with conventional side-grasp mug flipping, rim grasping provides a smaller contact area and tighter gripper tolerance.

\paragraph{Results.}

\begin{table}[!h]
\centering
\small
\begin{tabular}{l|ccc|c} \toprule
Tasks   & Pick and Place & Towel Folding & Rim-Grasp Cup Placement & Average \\ \midrule
\# of Demos   & 59 & 59 & 99 & -- \\ \midrule
DP      & 80    & 70   & 0 & 50.0           \\
UVA     & 0    & 30   & 0 & 10.0                 \\
$\pi_0$ & 0  & 20  & 0 & 6.7            \\ \midrule
DAP &  \textbf{90}  & \textbf{80} & \textbf{30} & \textbf{66.7}         \\
DAP (w/o Dyn.) & 40 & 40 & 10 & 30.0 \\\bottomrule
\end{tabular}
\vspace{5pt}
\caption{Real-world task success rates (\%) and dynamics-ablation results.}
\label{tab:main_real}
\end{table}

% \begin{table}[H]
% \centering
% \begin{tabular}{@{}c|ccc|c@{}}
% \toprule
% MSE $\downarrow$        & PP & TF & RCP & Average \\ \midrule
% \textbf{DP}   & \textbf{0.21} & \textbf{0.0125} & 0.637  & 0.287       \\
% \textbf{UVA}  & 2.16  & 0.803 & 1.82      & 1.59       \\
% \textbf{$\boldsymbol{\pi_0}$}   & 1.07  & 0.643 & 0.758 & 0.824     \\
% \textbf{DAP (Ours)} & 0.294  & 0.0652 & \textbf{0.323} & \textbf{0.227}  \\ \bottomrule
% \end{tabular}
% \vspace{+5pt}
% \caption{Action prediction error (MSE$\downarrow$) on validation dataset for each task. 
% %Each column indicates task name: CD (Close Drawer), PP (Pick and Place), TF (Towel Folding), and RCP (Rim-Grasp Cup Placement).
% Task abbreviations: PP (Pick and Place), TF (Towel Folding), RCP (Rim-Grasp Cup Placement).
% }
% \label{tab:val_mse}
% \vspace{-10pt}
% \end{table}

DAP achieves an average success rate of 66.7\% (Table~\ref{tab:main_real}), outperforming the strongest baseline, DP, at 50\%.
On the especially challenging RCP task, where rim grasping, reorientation, and stable placement make early contact choices critical, DAP achieves 3/10 successes while all non-DAP baselines achieve 0/10.
Removing dynamics alignment reduces the average success rate from 66.7\% to 30.0\%, a 55.0\% relative drop, and DP outperforms DAP (w/o Dyn.) on average.
Together with the simulation ablation in Table~\ref{tab:sim_dyn_ablation}, this indicates that the improvement is not due to the flow matching policy architecture alone, but to conditioning on the future predicted by the dynamics model.
Because $\pi_0$ and UVA target large-scale multi-task pretraining, their weak results under our far smaller demonstration budgets are expected rather than a tuning artifact, and Appendix~\ref{app:implementation_details} details each baseline's configuration. 
Our claim is not that DAP beats foundation-scale policies in general, but that dynamics-aligned refinement helps with a relatively modest number of demonstrations.

\section{Discussion}
\label{sec:discussion}

% We presented Dynamics-Aligned Flow Matching Policy, which improves policy generalization by incorporating explicit dynamics feedback during generation. 
% Our key insight is that flow extrapolation enables intermediate action flow samples to produce terminal action estimates, which then condition the dynamics model to provide corrective feedback to the policy. 
% The significant performance gain validates our core hypothesis that dynamics-aligned action generation enables more robust robotic policies. 
% The approach offers a practical solution for improving sample efficiency and robustness to future-prediction failures. \\
We presented DAP, an inference-time mechanism for coupling generative action sampling with action-conditioned visual dynamics. Flow extrapolation lets intermediate action samples condition the dynamics model, whose predicted futures then correct the policy while generation is still ongoing. 
Across simulation and real-world experiments, conditioning on a dynamics-predicted future improves action generation, and the future-perturbation analysis shows that in-loop co-sampling restores action--future consistency when that future is corrupted, which post-hoc or fixed future conditioning cannot. 
Because DAP is a sampling methodology rather than a specific dynamics-model architecture, its predictive module is replaceable, making the integration of recent world-model approaches~\cite{cosmos,vjpea2} a promising direction for future work.

\paragraph{Limitations.}
DAP assumes that the dynamics model can predict the observed scene reasonably well.
Additionally, the architecture relies on SD-VAE latents, whose photo-realistic prior causes a simulation-domain gap that degrades absolute performance.
Thus, VAE fine-tuning was necessary.
Our real-world evaluation uses 10 trials per task, so the per-task rates carry wide confidence intervals and should be read as indicative.
Our real-world comparison also focuses on the 50--100 demonstration regime, and larger training sets or more specialized fine-tuning may further improve video-action and pretrained VLA baselines such as UVA and $\pi_0$.
While our work demonstrates improvements in robustness within single-task, single-dataset scenarios, future work should explore extending DAP to generalization across different tasks and completely unseen object types using diverse datasets.

%===============================================================================

% The acknowledgments are automatically included only in the final and preprint versions of the paper.
\acknowledgments{
    This work is in part supported by the National Research Foundation of Korea (NRF, RS-2024-00451435(20\%), RS-2024-00413957(20\%)), Institute of Information \& communications Technology Planning \& Evaluation (IITP, RS-2025-02305453(15\%), RS-2025-02273157(15\%), RS-2025-25442149(15\%), RS-2021-II211343(15\%)) grant funded by the Ministry of Science and ICT (MSIT), Institute of New Media and Communications (INMAC), and the BK21 FOUR program of the Education, Artificial Intelligence Graduate School Program (Seoul National University), and Research Program for Future ICT Pioneers, Seoul National University in 2026.}

%===============================================================================

\clearpage
\bibliography{main}  % .bib
\clearpage

\appendix
\raggedbottom
\makeatletter\setlength{\@fptop}{0pt}\makeatother
\section*{Appendix}

\section{Additional Analysis}
\subsection{\texorpdfstring{$\Delta_\text{PSNR}$}{Delta PSNR} Action-Sensitivity Evaluation}
\label{app:delta_psnr}
Following the $\Delta_\text{PSNR}$ protocol introduced in Genie~\cite{genie}, we evaluate whether the learned dynamics model uses the action input rather than only generating visually plausible future frames. On PickCube-v1, we compare action-conditioned dynamics models with action-blind video models trained with $N\in\{100,200,500\}$ demonstrations and evaluate all checkpoints on the same disjoint held-out test set. For each sample, we compare future-observation PSNR under the ground-truth action against the mean PSNR obtained from 10 uniformly sampled random actions, defining $\Delta_\text{PSNR}$ as their difference. The video model is an action-blind control, so its $\Delta_\text{PSNR}$ should remain near zero if the metric isolates true action sensitivity.

Table~\ref{tab:psnr} reports paired bootstrap estimates with 95\% confidence intervals over $n=1600$ samples. The action-conditioned dynamics model becomes increasingly action-discriminative as the number of demonstrations grows: $\Delta_\text{PSNR}$ increases monotonically from $+4.75$ to $+5.09$ and $+6.29$~dB for 100, 200, and 500 demonstrations, respectively. In contrast, the video-only model has $\Delta_\text{PSNR}\approx 0$ at every scale, and all of its confidence intervals include zero, confirming that visually plausible next-frame prediction alone does not imply action conditioning. Although the video model eventually achieves higher PSNR under the true action at 500 demonstrations by learning a strong temporal prior, $\Delta_\text{PSNR}$ separates this static prediction quality from genuine action sensitivity.

\subsection{Flow Extrapolation Evaluations}
\label{app:flow_extrapolation_eval}
\textbf{Flow Extrapolation.}
For a rectified-flow linear-interpolant model, training pairs a clean sample
$x_1$ with Gaussian noise $x^{(0)}$ through
\begin{equation}
    x^{(\tau)} = (1-\tau)x^{(0)} + \tau x_1,
    \qquad \tau \in [0,1],
    \label{eq:appendix_flow_interpolant}
\end{equation}
with target velocity $u^{(\tau)} = x_1 - x^{(0)}$.
Given an intermediate sample $x^{(\tau)}$ and a predicted velocity $\hat{u}^{(\tau)}$,
the one-step extrapolation to the clean endpoint is
\begin{equation}
    \mathcal{E}(x^{(\tau)})
    = x^{(\tau)} + (1-\tau)\hat{u}^{(\tau)}(x^{(\tau)}).
    \label{eq:appendix_flow_extrapolation}
\end{equation}
This is the flow matching analogue of the diffusion ``$\hat{x}_0$''
reparameterization: rather than taking only a local Euler step
$x^{(\tau)} + d\tau\,\hat{u}^{(\tau)}$, it projects the current sample to
$\tau=1$ as if the velocity were constant over the remaining interval.
The resulting estimate is available at every intermediate flow step and is
exact only when the predicted velocity matches the true conditional velocity.

In DAP, the action flow and observation-latent flow are sampled jointly, and
each network conditions on the other flow's extrapolated endpoint rather than
on its noisy intermediate sample. Specifically, the policy receives the
extrapolated future observation $\mathcal{E}(o^{(\tau)})$, while the
dynamics model receives the extrapolated action $\mathcal{E}(a^{(\tau)})$.
These clean-sample-shaped conditioning signals are what keep the two flows
aligned throughout the coupled sampling process.

\textbf{Visualization Experiment.}
We evaluate how close the extrapolated estimates are to the ground-truth
next observation and action along the flow integration schedule.
The diagnostic uses the held-out real-robot Rim-Grasp Cup Placement validation split with two
camera views and a 7-D end-effector action, with the same preprocessing
as deployment: resize to $180\!\times\!240$ and center-crop to
$160\!\times\!160$.
For every validation sample, we run the full coupled sampling loop and, at each of the $K$ flow steps,
compute PSNR between $\mathcal{E}(o^{(\tau)})$ and the
ground-truth next observation for each camera, and MSE between $\mathcal{E}(a^{(\tau)})$
and the ground-truth action. The final prediction at $\tau=1$, which is
the value deployed by the policy, is evaluated separately as a consistency
check.

\begin{figure}
    \centering
    \includegraphics[width=\linewidth]{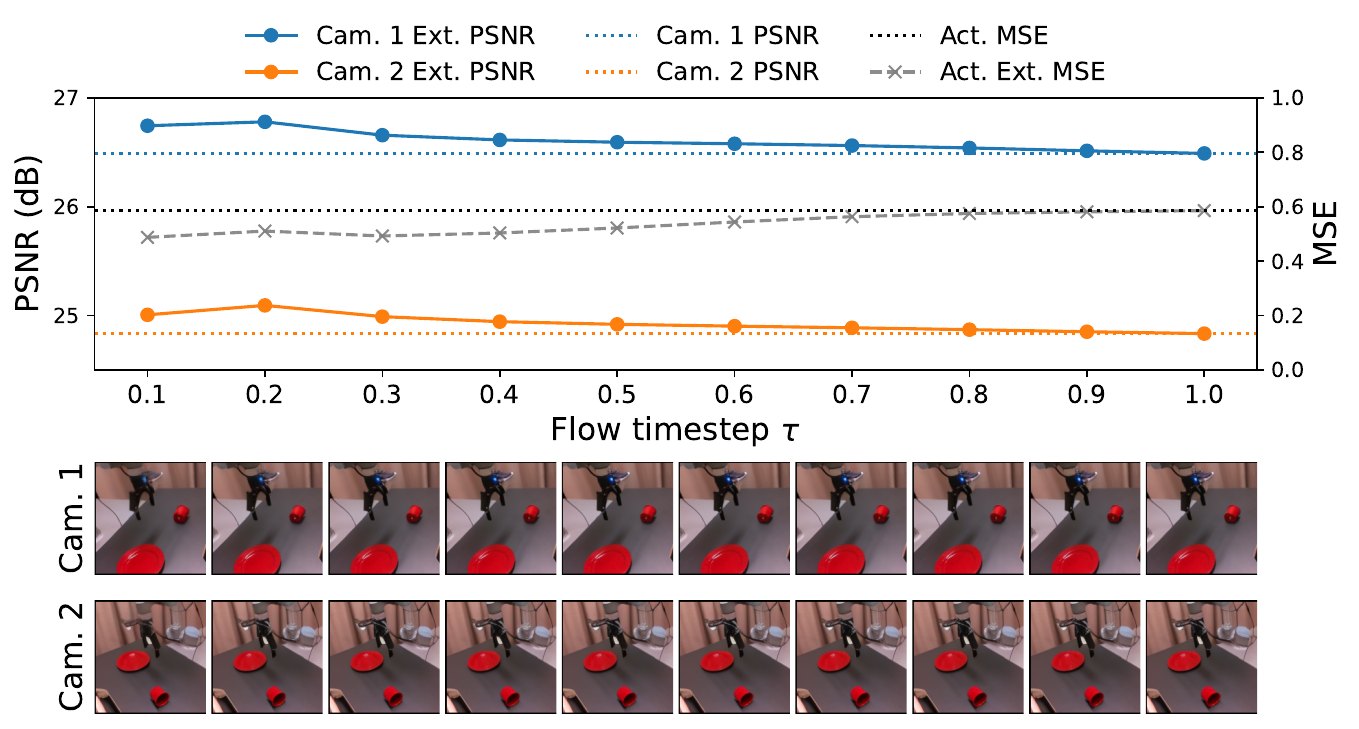}
    \caption{\textbf{Top:} Quantitative evaluation of extrapolated samples during flow generation. The peak signal-to-noise ratio (PSNR) values of the extrapolated future observation at each flow timestep $\tau$ are provided (Cam. 1 Ext. PSNR and Cam. 2 Ext. PSNR) with PSNR of final prediction (Cam. 1 PSNR and Cam. 2 PSNR). Likewise, we provide the mean squared error (MSE) values of extrapolated action for each $\tau$ (Act. Ext. MSE) with MSE of final prediction (Act. MSE). Evaluation is done on the entire validation dataset.
    \textbf{Bottom:} Qualitative samples of extrapolated future observation for each flow timestep $\tau$.
    }
    \label{fig:ext_figure}
    \vspace{-15pt}
\end{figure}

Figure~\ref{fig:ext_figure} shows two effects.
First, at early timesteps, the extrapolated endpoint already contains a coherent if blurry next frame, whereas the raw flow sample $o^{(\tau)}$ is still close to the initial noise when $\tau \approx 0$.
Second, the final per-step value matches the standalone final-step prediction, which confirms that the diagnostic measures the quantity used at deployment.
Together, these results justify cross-conditioning on
$(\mathcal{E}(o^{(\tau)}), \mathcal{E}(a^{(\tau)}))$ instead of the raw
$(o^{(\tau)}, a^{(\tau)})$.

% \begin{table}[!ht]
% \begin{tabular}{c|cccccccc|cc}\hline\hline
%      & \multicolumn{2}{c}{CD} & \multicolumn{2}{c}{PP} & \multicolumn{2}{c}{TF} & \multicolumn{2}{c}{RCP} & \multicolumn{2}{|c}{Average} \\ 
%      & Exp.          & Rand.         & Exp.          & Rand.          & Exp.          & Rand.          & Exp.          & Rand.           & Exp.          & Rand.       \\ \hline
% Ours & 25.85           & \textbf{23.64}          & 24.12            & \textbf{21.63}           & 24.67           & \textbf{22.76}           & 25.94            & \textbf{24.12}            & 25.15        & \textbf{23.04}        \\
% UVA  & \textbf{26.01}           & 20.29          & \textbf{24.57}            & 18.13           & \textbf{26.03}           & 20.11           & \textbf{26.51}            & 19.06            & \textbf{25.78}        & 19.40       \\ \hline\hline
% \end{tabular}
% \end{table}

\subsection{Effect of Random Data on Dynamics Prediction}

\begin{table}[!ht]
\centering
\begin{tabular}{c|cccccc|cc}\hline\hline
     & \multicolumn{2}{c}{PP} & \multicolumn{2}{c}{TF} & \multicolumn{2}{c}{RCP} & \multicolumn{2}{|c}{Average} \\ 
     & Exp.          & Rand.          & Exp.          & Rand.          & Exp.          & Rand.           & Exp.          & Rand.       \\ \hline
DAP & 24.12            & \textbf{21.63}           & 24.67           & \textbf{22.76}           & 25.94            & \textbf{24.12}            & 24.91        & \textbf{22.84}        \\
UVA  & \textbf{24.57}            & 18.13           & \textbf{26.03}           & 20.11           & \textbf{26.51}            & 19.06            & \textbf{25.70}        & 19.10       \\ \hline\hline
\end{tabular}
\vspace{10pt}
\caption{Per-task PSNR comparison between DAP and UVA on real-world data, reported separately on expert trajectories (Exp.) and random trajectories (Rand.) for Pick and Place (PP), Towel Folding (TF), Rim-Grasp Cup Placement (RCP), and their average.}
\label{tab:psnr_real}
\end{table}

Inspired by using random data for dynamics modeling~\cite{causalpik} and leveraging randomness for generalization~\cite{demodice,edac,spqr}, we train the dynamics model $p(o_{t+1}|o_t,a_t)$ with both expert and random data to expand observation-action coverage.
Table~\ref{tab:psnr_real} compares real-world future-observation PSNR for DAP and UVA on validation datasets containing both expert demonstrations and random trajectories.
UVA is slightly better on expert trajectories, which is expected because it trains in both video-prediction and forward-dynamics modes during policy learning, effectively increasing its dynamics-learning duration relative to DAP.
However, UVA's prediction quality degrades substantially on random trajectories, whereas DAP maintains more consistent performance.
This contrast suggests that DAP learns a more generalizable action-conditioned dynamics model, while UVA is more specialized to expert visual patterns.

% We conjecture that such balanced dynamics learning contributes to DAP's superior OOD generalization.
% By maintaining consistent prediction quality across diverse trajectories, our model develops a more robust understanding of environment dynamics that transfers better to novel scenarios.
% This hypothesis aligns with our experimental results, where DAP demonstrates improved robustness particularly in challenging OOD conditions.

% \begin{figure}[!h]
%     \centering
%     \includegraphics[width=\textwidth]{img/psnr_real_task_avg_merged.pdf}
%     \caption{UVA and DAP}
%     \label{fig:example_image}
% \end{figure}

\subsection{Inference Time}
\label{app:inference_time}
We compare end-to-end action-generation latency and the additional cost introduced by dynamics alignment to evaluate practical deployability in closed-loop control.
Low latency is important because the policy repeatedly replans during execution, leaving only a limited computation budget per control cycle.
Table~\ref{tab:infer_time} reports the mean and standard deviation over 100 trials.
For DAP, this measurement includes the full policy--dynamics co-sampling procedure used at test time; for DP, it uses the ManiSkill evaluation implementation from our experiments.

\Needspace{9\baselineskip}
\begin{center}
\begin{minipage}{\linewidth}
\centering
\small
\captionof{table}{End-to-end inference latency (mean $\pm$ standard deviation, milliseconds) over 100 trials.}
\label{tab:infer_time}
\begin{tabular}{@{}lcc@{}}
\toprule
Method & DP & DAP \\
\midrule
Latency & $699.27 \pm 3.01$ & $327.03 \pm 8.34$ \\
\bottomrule
\end{tabular}
\end{minipage}
\end{center}

Although DAP uses an additional action-conditioned dynamics model during sampling, it remains more than twice as fast as DP.
This latency gap is primarily attributable to the sampler design: DP uses DDPM-style iterative denoising, whereas DAP uses flow matching with fewer integration steps.
This suggests that iterative dynamics feedback does not preclude real-time use.
The model size and offline-training overheads are also modest: DAP's total trainable parameter count, including the dynamics model, is comparable to that of DP, and the dynamics model converges in under 100 epochs on all real-world datasets, including those with random trajectories.
Thus, the added dynamics component imposes limited computational overhead relative to the primary policy optimization while providing the self-correction mechanism evaluated above.

\subsection{Does DAP Capture Expert Action Multimodality?}
\label{app:bic}
\label{sec:bic_main}

We also test whether mutual policy--dynamics correction prevents
collapse to a single high-likelihood action mode when the local expert behavior
is multimodal, by measuring whether each policy's per-state action
samples preserve the same local mode-count structure as expert demonstrations.
A small $D_\text{JS}(P_\pi \| P_\text{exp})$ indicates that the policy $\pi$
produces sample clouds whose BIC-selected mode counts track the per-state
mode-count distribution of the local expert manifold. This does \emph{not} claim
that the policy reproduces the exact expert conditional at any individual state;
rather, it tests whether the policy's diversity, aggregated across states,
matches the diversity exhibited by nearby expert trajectories. As shown in
Table~\ref{tab:bic_jsd}, DAP achieves lower JS divergence to the expert than DP
on all three real-world tasks, indicating that DAP's per-state action samples
track the local structure of the expert demonstration manifold more faithfully
than the baseline. This result is consistent with the manifold-adherence view
recently advanced for generative control policies~\citep{pan2025muchado}.

\begin{center}
\begin{minipage}{\linewidth}
\centering
\small
\captionof{table}{Jensen--Shannon divergence between policy and expert BIC-optimal mode-count distributions (lower is closer to expert). DAP reduces the gap to the expert distribution on every task.}
\label{tab:bic_jsd}
\begin{tabular}{lcc}
\toprule
JSD $\downarrow$ & DP & DAP \\
\midrule
PP     & 0.128 & \textbf{0.086} \\
TF     & 0.165 & \textbf{0.092} \\
RCP    & 0.142 & \textbf{0.125} \\
\bottomrule
\end{tabular}
\end{minipage}
\end{center}

\paragraph{Compute Details.}
For each task, we pool $S{=}500$ states uniformly across the full dataset. For
each state $o_s$, the expert sample set contains the action chunks of the
$K{=}20$ nearest neighbors of $o_s$ in proprioceptive end-effector space, which
characterizes the \emph{local expert manifold} around $o_s$. DP and DAP are each
evaluated using $N{=}100$ independent action chunks sampled from the policy
conditioned on $o_s$, characterizing each policy's conditional distribution at
that state. We intentionally do not match $K$ and $N$: enlarging $K$ would draw
expert chunks from increasingly distant states and dilute the local expert
measurement, whereas policy samples are genuine i.i.d. conditional samples.

Each action chunk $a \in \mathbb{R}^{H\times 7}$ is reduced to its chunk-mean
rotation $r_{\mathrm{mean}}(a)=\frac{1}{H}\sum_{t=1}^{H}a_t[3{:}6]$, a 3-D feature that
captures the dominant variation in end-effector reorientation. For each source
$\mathcal{S}\in\{\text{exp},\text{DP},\text{DAP}\}$ and each state, we fit GMMs
with $k\in\{1,2,3,4,5\}$ full-covariance components to the resulting point cloud
and select the component count by the Bayesian Information Criterion
(BIC)~\citep{schwarz1978bic}:
\begin{equation}
    \mathrm{BIC}(k) \;=\; -2\log\hat{L}_k + p_k\,\log n,
    \label{eq:bic}
\end{equation}
where $\hat{L}_k$ is the maximized likelihood under the $k$-component GMM,
$p_k$ is the number of free parameters, and $n$ is the number of points in the cloud. The optimal count is
$k^\star=\arg\min_{k\in\{1,\dots,5\}}\mathrm{BIC}(k)$. Mixture models are
standard in robot learning for estimating behavioral components from
demonstrations~\citep{calinon2019gaussian,jankowski2025midigap}. We use the
\texttt{scikit-learn} implementation with default initialization and 100 EM
iterations.

Aggregating selected counts across states gives an empirical mode-count
distribution for each source,
\begin{equation}
    P_\mathcal{S}(k) \;=\; \frac{1}{S}\sum_{s=1}^{S}\mathds{1}[k^\star_\mathcal{S}(s) = k],
    \quad k\in\{1,\dots,5\}.
\end{equation}
We compare each policy distribution to the expert distribution using
Jensen--Shannon divergence,
\begin{equation}
    D_\text{JS}(P\,\|\,Q) \;=\; \tfrac{1}{2}D_\text{KL}\!\left(P \,\|\, M\right)
                          \;+\; \tfrac{1}{2}D_\text{KL}\!\left(Q \,\|\, M\right),
    \quad M = \tfrac{1}{2}(P+Q),
    \label{eq:jsd}
\end{equation}
which is symmetric, bounded in $[0,\log 2]$, and commonly used to compare expert
and learner action distributions in imitation
learning~\citep{ho2016gail,ke2020fdiv,ghasemipour2020divergence}. We apply
Laplace smoothing with pseudocount $1$ to avoid undefined KL terms when a bin is
empty. As a robustness check, we repeated the analysis with $\mathit{diagonal}$
and $\mathit{spherical}$ covariance. Spherical covariance inflates mode counts
uniformly across sources because a single elongated cloud must be tiled by
multiple isotropic components, but the relative ranking between
$D_\text{JS}(P_\text{DAP}\,\|\,P_\text{exp})$ and
$D_\text{JS}(P_\text{DP}\,\|\,P_\text{exp})$ is preserved on all three tasks.

\subsection{Classifier-Free Guidance Scale Sweep}
\label{app:cfg_sweep}
We additionally sweep the classifier-free guidance (CFG) scale on PickCube-v1 for both the dynamics model and the policy model. Because DAP uses two coupled conditional generators, we tune the two guidance mechanisms separately. For the dynamics model, which predicts the next-observation latent from the current observation and candidate action, we apply CFG only to the action condition $\hat{a}=\mathcal{E}(a^{(\tau)})$ while keeping the current observation as standard conditioning:
\begin{equation}
    \hat{u}_{o,\mathrm{CFG}}^{(\tau)}
    = \hat{u}_{o|o_t}^{(\tau)}
    + s_{a}\bigl(\hat{u}_{o|o_t,\hat{a}}^{(\tau)} - \hat{u}_{o|o_t}^{(\tau)}\bigr),
    \label{eq:cfg_dynamics}
\end{equation}
where $\hat{u}_{o|o_t}^{(\tau)}$ is conditioned only on the current observation, and $\hat{u}_{o|o_t,\hat{a}}^{(\tau)}$ is conditioned on both the current observation and the candidate action. The dynamics sweep varies only the action-guidance scale $s_a$. For the policy model, we likewise apply CFG only to the predicted future-observation condition $\hat{o}=\mathcal{E}(o^{(\tau)})$ while keeping the current observation as standard conditioning:
\begin{equation}
    \hat{u}_{a,\mathrm{CFG}}^{(\tau)}
    = \hat{u}_{a|o_t}^{(\tau)}
    + s_{o}\bigl(\hat{u}_{a|o_t,\hat{o}}^{(\tau)} - \hat{u}_{a|o_t}^{(\tau)}\bigr),
    \label{eq:cfg_policy}
\end{equation}
where $\hat{u}_{a|o_t}^{(\tau)}$ is conditioned only on the current observation, and $\hat{u}_{a|o_t,\hat{o}}^{(\tau)}$ is conditioned on both the current observation and the predicted future observation.

\begin{figure}[!h]
    \centering
    \includegraphics[width=\linewidth]{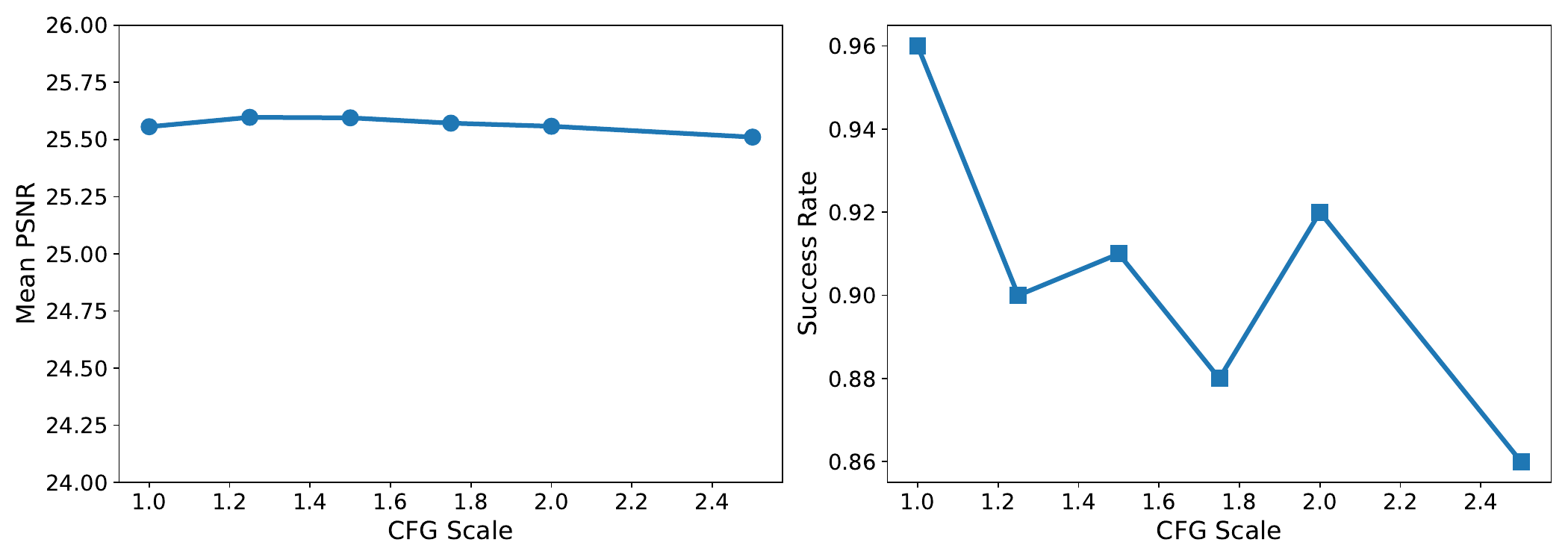}
    \caption{Mean future-observation PSNR and success rate of the PickCube-v1 DAP policy as a function of classifier-free guidance (CFG) scale. Left: dynamics action-guidance sweep. Right: policy future-observation-guidance sweep.}
    \label{fig:cfg_sweep_dyn_policy}
\end{figure}

Figure~\ref{fig:cfg_sweep_dyn_policy} visualizes the dynamics PSNR sweep and the policy success-rate sweep for PickCube-v1. The dynamics model is largely insensitive to the action CFG scale: mean PSNR varies by less than $0.1$~dB across $s_a\in\{1.0,1.25,1.5,1.75,2.0,2.5\}$, with the best value at $s_a=1.25$ only marginally above the default conditional setting. Thus, stronger action guidance does not materially improve future-observation fidelity on this task.

For the policy, the default future-observation guidance scale performs best: DAP achieves 96\% success at $s_{o}=1.0$. Increasing $s_{o}$ does not improve performance, but the policy remains robust over the tested range, with success rates of 90\%, 91\%, 88\%, 92\%, and 86\% for $s_{o}=1.25$, $1.5$, $1.75$, $2.0$, and $2.5$, respectively. We therefore use the default conditional scale $s_a=s_{o}=1.0$ unless otherwise stated; the sweep suggests that DAP's gains do not rely on aggressive CFG tuning.

\subsection{Task Complexity Analysis}
Among the three real-world tasks, Rim-Grasp Cup Placement is the most challenging: baseline methods consistently fail on it, while DAP still achieves non-trivial success. To characterize this gap, we analyze expert demonstrations with two complementary rotation statistics. Diffusion Policy also identifies complex 3D reorientation as a key challenge for visuomotor policies~\cite{DP}; accordingly, we examine both the total reorientation within each trajectory and the range of end-effector orientations visited across the dataset.

Specifically, we compute \emph{cumulative rotation} as the sum of geodesic angle changes between consecutive end-effector orientations. We also analyze the absolute end-effector rotation angle over all timesteps. The first statistic measures how much reorientation an episode requires, while the second measures how wide a rotational action space the task occupies.

\begin{figure}[!h]
    \centering
    \includegraphics[width=\linewidth]{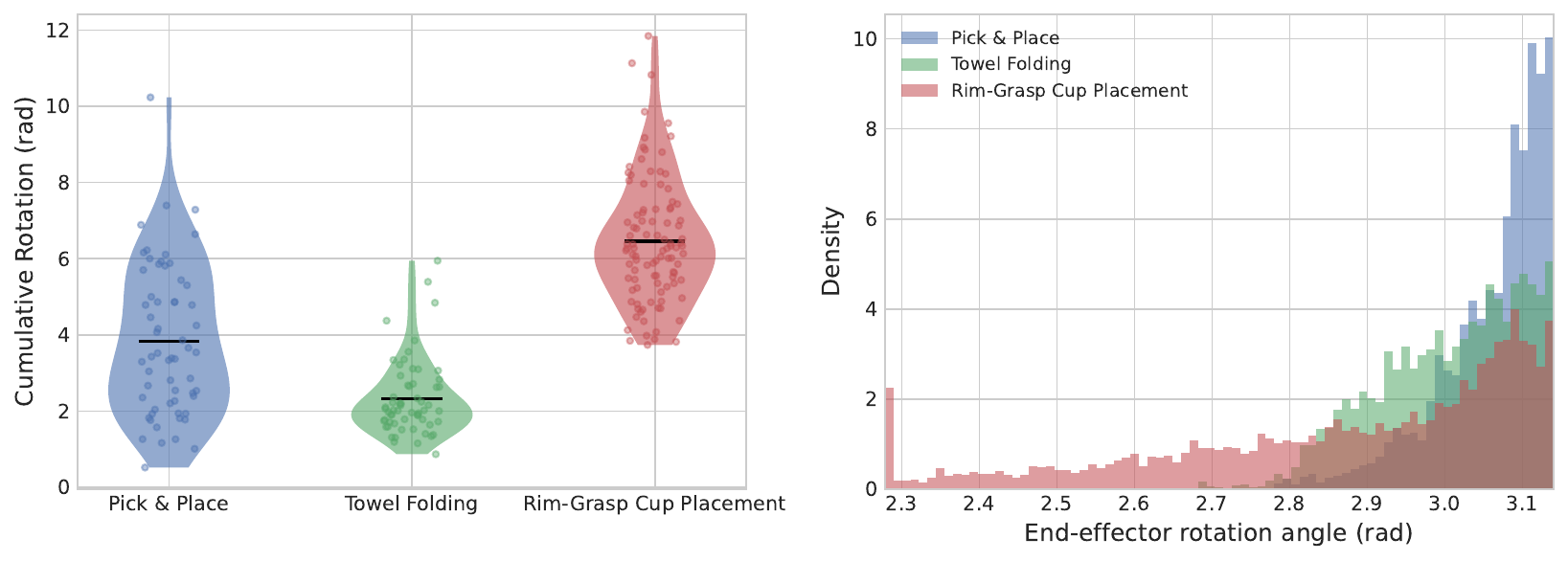}
    \caption{Rotation-based task-difficulty summary for Pick and Place, Towel Folding, and Rim-Grasp Cup Placement. Left: per-episode cumulative end-effector rotation shown as violin plots with jittered episode samples and mean markers. Cumulative rotation is computed as the sum of geodesic angle changes between consecutive end-effector orientations. Right: density-normalized histogram of absolute end-effector rotation angles aggregated over all timesteps.}
    \label{fig:task_difficulty_summary}
\end{figure}

Figure~\ref{fig:task_difficulty_summary} shows that Rim-Grasp Cup Placement is difficult for two reasons. First, its cumulative-rotation distribution is substantially higher than those of Pick and Place and Towel Folding, indicating longer and more sustained reorientation. Second, it has the broadest absolute-angle distribution, indicating a wider rotational action space. In practice, the policy must grasp the rim, rotate the cup upright, and refine placement across a wider range of wrist poses, making the task more multimodal and less tolerant to small visual or contact errors~\cite{DP}.

\subsection{Train--Test Conditioning Shift}
\label{app:train_test_shift}
The policy is trained on ground-truth futures but conditioned on DAP's predicted futures at test time.
On 300 held-out PickCube-v1 states, replacing the ground-truth future with DAP's prediction changes the executed action by only $8.0\times10^{-4}$ per element, and the action error against the expert stays comparable at $4.77\times10^{-3}$ with the prediction versus $4.57\times10^{-3}$ with the ground truth, so DAP's prediction can stand in for the ground truth.
A corrupted fixed future, in contrast, lowers success (Figure~\ref{fig:ablation}(a)).

\clearpage
\newpage

\section{Implementation Details}
\label{app:implementation_details}
\subsection{Task Details}
\subsubsection{Simulation Tasks}
We evaluate in simulation on four ManiSkill tasks: PickCube-v1, PushCube-v1, DrawTriangle-v1, and StackCube-v1. For each task, we run 100 evaluation rollouts and report success rate as the percentage of episodes that satisfy the environment success condition within the rollout horizon. For fair comparison, all methods use the same maximum of 300 episode steps on every task. Table~\ref{tab:sim_tasks} summarizes the task definitions, success criteria, evaluation horizons, and task interfaces.

\begin{table}[ht]
\centering
\caption{Simulation task definitions, success criteria, evaluation horizons, and task interfaces. All methods are evaluated with the same maximum episode steps for each task.}
\label{tab:sim_tasks}
\footnotesize
\setlength{\tabcolsep}{4pt}
\resizebox{\textwidth}{!}{%
\begin{tabular}{@{}p{0.14\linewidth}p{0.23\linewidth}p{0.35\linewidth}ccp{0.12\linewidth}@{}}
\toprule
Task & Description & Success criterion & Action dim & Max steps & Camera views \\
\midrule
\textbf{PickCube-v1} & Grasp-and-place task that moves a cube to a target goal position. & The cube position is within \texttt{goal\_thresh} of the goal position (default Euclidean threshold: 0.025m). & 7 & 300 & base + hand \\
\textbf{PushCube-v1} & Non-prehensile pushing task that moves a cube to a goal region in front of the robot. & The cube's xy position is within \texttt{goal\_radius} of the target xy position (default Euclidean threshold: 0.1m), and the cube remains on the table. & 4 & 300 & base + hand \\
\textbf{DrawTriangle-v1} & Drawing task that traces a goal triangle on a canvas using a stick. & The drawn points are within a Euclidean distance of 0.05m from points on the goal triangle. & 3 & 300 & base \\
\textbf{StackCube-v1} & Pick up a red cube, stack it on top of a green cube, and release it without the cube falling. Both cubes have randomized z-axis rotations and non-colliding xy positions on the table. & The red cube is on top of the green cube within half a cube size, is static, and is not being grasped by the robot. & 7 & 300 & base + hand \\
\bottomrule
\end{tabular}%
}
\end{table}

\subsubsection{Real-World Tasks}
We conduct real-world experiments on a UR5 robotic arm with teleoperation via a 3DConnexion SpaceMouse. 
The workspace is monitored by three Intel RealSense D435 cameras, two exocentric cameras and one egocentric. 
For each task, we collect 50--100 expert demonstration episodes through teleoperation, with the number varying by task complexity. 
We also collect random trajectory datasets of similar size to enable comprehensive dynamics learning.

For real-world experiments, each method is tested across 10 trials per task with different initial configurations. We report success rate as the percentage of trials completed within a 60--90 second time limit. All policies are trained on identical expert demonstrations and evaluated under the same environmental configurations.

We evaluate on three real-world manipulation tasks. \textbf{Pick and Place} (PP) requires picking an object from a random initial pose and placing it at the designated target. \textbf{Towel Folding} (TF) is a deformable-object task in which the robot grasps a towel edge and folds it in half. \textbf{Rim-Grasp Cup Placement} (RCP) requires grasping a horizontally placed cup by its rim and placing it upright on a plate; success is defined by stable placement in the target region. Compared with conventional mug-flipping settings that allow grasping the side surface, our Rim-Grasp Cup Placement task requires rim grasping with a smaller contact area and tighter pose tolerance.

\subsection{DAP}
\begin{algorithm}[t]
    \caption{Sampling with flow extrapolation}
    \begin{algorithmic}[1]
        \State Given $o_t$, set $d\tau = 1 / K$
        \State Sample $a^{(0)} \sim \mathcal{N}(0, I), \quad o^{(0)} \sim \mathcal{N}(0, I)$
        \State Initialize $\hat{a} \leftarrow a^{(0)}+\pi_\phi(a^{(0)}, 0 | o_t, \varnothing), \quad \hat{o} \leftarrow o^{(0)}+f_\theta(o^{(0)}, 0 | o_t, \varnothing)$
        \For{$\tau$ in $0, d\tau, \dots, 1-d\tau$}
            \State $\hat{u}_a^{(\tau)} \leftarrow \pi_\phi(a^{(\tau)}, \tau | o_t, \hat{o})$
            \State $\hat{u}_o^{(\tau)} \leftarrow f_\theta(o^{(\tau)}, \tau | o_t, \hat{a})$
            \State $a^{(\tau+d\tau)} \leftarrow a^{(\tau)} + d\tau \cdot \hat{u}_a^{(\tau)}$
            \State $o^{(\tau+d\tau)} \leftarrow o^{(\tau)} + d\tau \cdot \hat{u}_o^{(\tau)}$
            \State $\hat{a} \leftarrow a^{(\tau)} + (1-\tau)\hat{u}_a^{(\tau)}, \quad \hat{o} \leftarrow o^{(\tau)} + (1-\tau)\hat{u}_o^{(\tau)}$
        \EndFor
        \State $\hat{o}^{\DAP} \leftarrow o^{(1)}, \hat{a}^{\DAP} \leftarrow a^{(1)}$
        \State \Return $\hat{o}^{\DAP}, \hat{a}^{\DAP}$
    \end{algorithmic}
    \label{alg:sample}
\end{algorithm}

\begin{figure*}[!t]
% \vskip 0.2in
%\vspace{-10pt}
    \centering
    \includegraphics[width=\textwidth]{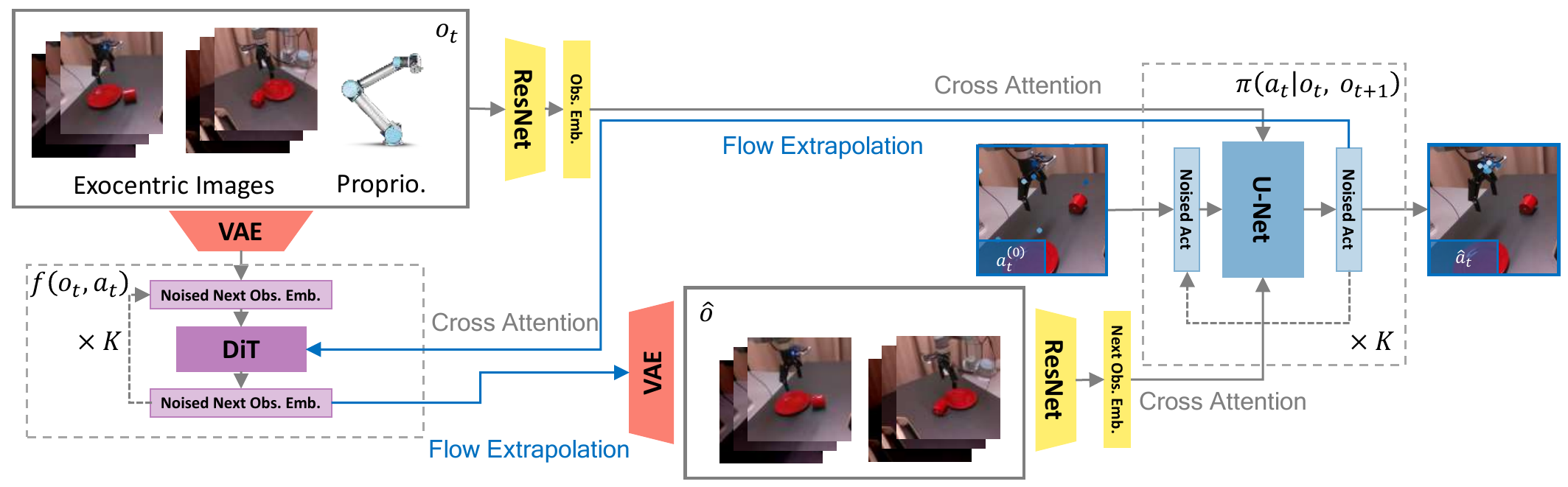}
    \caption{The dynamics model $f(o_t,a_t)$ (left) learns from mixed expert and random data using a flow matching VAE-DiT architecture with cross-attention conditioning. 
    The policy $\pi(a_t|o_t,o_{t+1})$ generates actions through flow matching with a cross-attention U-Net architecture. 
    Flow extrapolation (blue arrows) enables mutual correction: it maps flow samples at timestep $\tau$ to endpoint estimates for $o_{t+1}$ and $a_t$, aligning action generation with dynamics predictions.
    This iterative parallel generation ensures consistent and robust behaviors while maintaining real-time performance.}
    \label{fig:arch}
    \vspace{-10pt}
\end{figure*}

% \textbf{Dynamics Model.} For the dynamics model, we leverage the DiT architecture following Cosmos \cite{cosmos}.
% Instead of the video tokenizer in the original implementation, we use the pretrained VAE of Stable Diffusion \cite{SD1} for latent diffusion. 
% In addition, the action is conditioned on the dynamics model through cross-attention layers.
% %rather than being added to the diffusion timestep.
% The images are randomly cropped in the training stage and center-cropped in the inference stage.

% \textbf{Policy Model.} For the policy model, we mostly follow the same architecture and training details as diffusion policy.
% To effectively inject the observations into the network, we introduce cross-attention layers between layers of U-Net used by Stable Diffusion~\cite{SD1}.
% We separate the vision encoder of $o_t$ from that of $o_{t+1}$ and train them independently. We provide two exocentric views as observations, coincident with the dynamics model's prediction.

For DAP, we use a consistent two-stage training protocol across all four ManiSkill tasks. 
We first train the action-conditioned dynamics model for 100 epochs, and then train the policy for 500 epochs using only expert demonstrations. 
The policy also conditions on past actions through an auxiliary pathway that is separate from the future-observation conditioning and is not updated during co-sampling.
It is implemented with cross-attention or FiLM depending on the task (Table~\ref{tab:dap_hparams_eval_settings}).
For details on the sampling algorithm and architecture design, see Algorithm~\ref{alg:sample} and Figure~\ref{fig:arch}.
%During policy training, we evaluate checkpoints every 10 epochs and report the full checkpoint trajectory in Table~\ref{tab:dap_success_rate_by_epoch}. 
%The best checkpoints for PickCube-v1, PushCube-v1, and DrawTriangle-v1 occur at 100 epochs, 30 epochs, and 290 epochs, respectively, indicating that the simpler tasks converge earlier whereas DrawTriangle-v1 benefits from substantially longer optimization.

To make the comparison with UVA as fair as possible, we intentionally keep the DAP optimization setup almost identical to the UVA setup whenever the architecture allows it. In particular, both methods use the same two-stage budget of 100/500 epochs, the same learning-rate scale, the same beta values, cosine scheduling, 1000 warmup steps, and the same EMA configuration. The main differences are architecture-specific choices required by DAP, including the shorter observation/action horizon, different batch size, and numerical precision. 

  \begin{table}[h]
  \centering
  \caption{Hyperparameters and evaluation settings used for DAP on PickCube-v1, PushCube-v1, DrawTriangle-v1, and StackCube-v1. All values are shared across the four tasks except the past-action conditioning pathway.}
  \label{tab:dap_hparams_eval_settings}
  \resizebox{\textwidth}{!}{%
  \begin{tabular}{ll|cccc}
  \toprule
   &  & PickCube-v1 & PushCube-v1 & DrawTriangle-v1 & StackCube-v1 \\
  \midrule
  \multicolumn{6}{l}{\textbf{Hyperparameters}} \\
   & batch\_size               & \multicolumn{4}{|c}{64} \\
   & policy\_lr\_peak          & \multicolumn{4}{|c}{1.00e-04} \\
   & policy\_betas             & \multicolumn{4}{|c}{(0.9, 0.95)} \\
   & policy\_weight\_decay     & \multicolumn{4}{|c}{0.02 (selective)} \\
   & dynamics\_lr\_peak        & \multicolumn{4}{|c}{1.00e-04} \\
   & dynamics\_betas           & \multicolumn{4}{|c}{(0.9, 0.95)} \\
   & dynamics\_weight\_decay   & \multicolumn{4}{|c}{1.00e-03} \\
   & lr\_scheduler             & \multicolumn{4}{|c}{cosine} \\
   & warmup\_steps             & \multicolumn{4}{|c}{1000} \\
   & num\_epochs (dynamics/policy)   & \multicolumn{4}{|c}{100/500} \\
   & mixed\_precision          & \multicolumn{4}{|c}{bf16} \\
   & max\_grad\_norm           & \multicolumn{4}{|c}{1.0} \\
   & EMA power / max           & \multicolumn{4}{|c}{0.75 / 0.9999} \\
   & horizon                   & \multicolumn{4}{|c}{16} \\
   & n\_obs\_steps             & \multicolumn{4}{|c}{2} \\
   & n\_flow\_steps            & \multicolumn{4}{|c}{10} \\
   & FM\_timestep\_sampling    & \multicolumn{4}{|c}{logit-normal} \\
   & future\_cond\_dropout     & \multicolumn{4}{|c}{0.1} \\
   & past\_action\_conditioning & cross-attn & FiLM & FiLM & cross-attn \\
  \midrule
  \multicolumn{6}{l}{\textbf{Evaluation Setting}} \\
   & n\_action\_steps   & \multicolumn{4}{|c}{8} \\
  \bottomrule
  \end{tabular}
  }
  \end{table}

\subsection{DP}
For DP in simulation, we use the official ManiSkill3 diffusion policy baseline implementation.\footnote{\url{https://github.com/haosulab/ManiSkill}}
For real-world experiments, we use the official Diffusion Policy implementation.\footnote{\url{https://github.com/real-stanford/diffusion_policy}}
We train and evaluate DP under the same demonstration budget as DAP; in simulation, we report the best checkpoint and the mean over the last three checkpoints (Table~\ref{tab:sim_main}), while in real-world experiments, we report the latest checkpoint and verify the implementation on a simple validation task.
The implementation details below and in Table~\ref{tab:dp_maniskill_hparams_eval_settings} are for the simulation setting.
The policy is \texttt{BC\_Diffusion\_rgbd\_UNet}: a DDPM action-sequence policy with a \texttt{PlainConv} RGB encoder, 256-dimensional visual features, and \texttt{ConditionalUnet1D(down\_dims=[64,128,256], n\_groups=8)}.
The DDPM scheduler uses 100 timesteps, a squared-cosine beta schedule, clipped samples, and $\epsilon$ prediction; predicted actions are clipped to $[-1,1]$.
\begin{table}[t]
\centering
\caption{Hyperparameters and evaluation settings used for the ManiSkill DP baseline on PickCube-v1, PushCube-v1, DrawTriangle-v1, and StackCube-v1. All values are shared across the four tasks.}
\label{tab:dp_maniskill_hparams_eval_settings}
\resizebox{\textwidth}{!}{%
\begin{tabular}{ll|cccc}
\toprule
 &  & PickCube-v1 & PushCube-v1 & DrawTriangle-v1 & StackCube-v1 \\
\midrule
\multicolumn{6}{l}{\textbf{Hyperparameters}} \\
 & batch\_size             & \multicolumn{4}{|c}{256} \\
 & lr\_peak                & \multicolumn{4}{|c}{1.00e-04} \\
 & betas                   & \multicolumn{4}{|c}{(0.95, 0.999)} \\
 & weight\_decay           & \multicolumn{4}{|c}{1.00e-06} \\
 & lr\_scheduler           & \multicolumn{4}{|c}{cosine} \\
 & warmup\_steps           & \multicolumn{4}{|c}{500} \\
 & total\_iters            & \multicolumn{4}{|c}{100k} \\
 & EMA power               & \multicolumn{4}{|c}{0.75} \\
 & horizon                 & \multicolumn{4}{|c}{16} \\
 & n\_obs\_steps           & \multicolumn{4}{|c}{2} \\
 & n\_action\_steps        & \multicolumn{4}{|c}{8} \\
 & n\_diffusion\_steps     & \multicolumn{4}{|c}{100} \\
\midrule
\multicolumn{6}{l}{\textbf{Evaluation Setting}} \\
 & eval\_envs              & \multicolumn{4}{|c}{10} \\
 & eval\_episodes          & \multicolumn{4}{|c}{100} \\
 & eval\_freq              & \multicolumn{4}{|c}{5k iters} \\
 & checkpoint              & \multicolumn{4}{|c}{best only} \\
\bottomrule
\end{tabular}%
}
\end{table}

\subsection{LPB}
For LPB, we use the official Latent Policy Barrier codebase\footnote{\url{https://github.com/zhanyisun/lpb}} and instantiate the baseline~\cite{lpb} as a three-stage ManiSkill3 pipeline: train a base DDPM image policy, train a latent visual dynamics model initialized from the policy encoder, and apply classifier-style barrier guidance during diffusion sampling.
Following the official protocol, we train the base policy for the same number of epochs as our policy, select the best checkpoint, estimate the barrier threshold from additional rollouts, and train the dynamics model for the same number of epochs as ours.
The base policy uses \texttt{DiffusionUnetHybridImagePolicy} with a ResNet-18 GroupNorm encoder, DDPM sampling with 100 diffusion steps, and U-Net down dimensions [512, 1024, 2048].
The latent dynamics model freezes the policy visual encoder and trains a ViT predictor with depth 6, 16 heads, MLP dimension 2048, dropout 0.1, action embedding dimension 240, proprioception embedding dimension 16, and frameskip 15.

We calibrate the barrier threshold $\tau$ without test-set access.
After checkpoint selection, we run the selected policy with LPB disabled by setting \texttt{threshold}$=10^9$ on 50 calibration seeds, record the current cost $\delta$ at every policy call, and set $\tau=\mathrm{median}(\delta)$.
During evaluation, LPB computes $\delta$ once per policy call; if $\delta>\tau$ and the reverse-diffusion timestep is below the guidance-start timestep, it applies the dynamics-model loss gradient to the current noisy action sample with guidance scale 0.05 (Table~\ref{tab:lpb_maniskill_hparams_eval_settings}).

\begin{table}[t]
\centering
\caption{Hyperparameters and evaluation settings used for the LPB baseline on PickCube-v1, PushCube-v1, DrawTriangle-v1, and StackCube-v1. Values are shared across tasks except for the calibrated threshold and resulting LPB fire rate.}
\label{tab:lpb_maniskill_hparams_eval_settings}
\resizebox{\textwidth}{!}{%
\begin{tabular}{ll|cccc}
\toprule
 &  & PickCube-v1 & PushCube-v1 & DrawTriangle-v1 & StackCube-v1 \\
\midrule
\multicolumn{6}{l}{\textbf{Hyperparameters}} \\
 & policy batch\_size        & \multicolumn{4}{|c}{64} \\
 & policy num\_epochs        & \multicolumn{4}{|c}{500} \\
 & dynamics batch\_size      & \multicolumn{4}{|c}{96} \\
 & dynamics num\_epochs      & \multicolumn{4}{|c}{100} \\
 & horizon                   & \multicolumn{4}{|c}{16} \\
 & n\_obs\_steps             & \multicolumn{4}{|c}{2} \\
 & n\_action\_steps          & \multicolumn{4}{|c}{8} \\
 & n\_diffusion\_steps       & \multicolumn{4}{|c}{100} \\
 & guidance\_start\_timestep & \multicolumn{4}{|c}{10} \\
 & guidance\_scale           & \multicolumn{4}{|c}{0.05} \\
\midrule
\multicolumn{6}{l}{\textbf{Evaluation Setting}} \\
 & checkpoint                & \multicolumn{4}{|c}{best validation} \\
 & threshold\_rule           & \multicolumn{4}{|c}{median $\delta$} \\
 & threshold $\tau$          & 1.367 & 0.808 & 1.664 & 4.636 \\
 & LPB fire\_rate            & 48.7\% & 50.3\% & 52.0\% & 51.2\% \\
\bottomrule
\end{tabular}%
}
\end{table}

\subsection{UVA}
For UVA, we use the official Unified Video Action implementation.\footnote{\url{https://github.com/ShuangLI59/unified_video_action}}
UVA jointly models videos and actions through masked training across video-generation, dynamics-prediction, and policy modes.
To align the baseline with DAP, we first train the video model for the same number of epochs as our dynamics model (100 epochs) and then train the policy in unified mode for the same number of epochs as our policy model (500 epochs). We evaluate every 10 epochs during policy training. UVA conditions on four history frames, subsampled with a stride of 4 from the most recent 16 observation steps (n\_obs\_steps $=16$ in Table~\ref{tab:uva_maniskill_hparams_eval_settings}), and predicts four future frames at the same stride, whereas DP and DAP use two frames. 

\begin{table}[t]
  \centering
  \caption{Hyperparameters and evaluation settings used for UVA on PickCube-v1, PushCube-v1, DrawTriangle-v1, and StackCube-v1. All values are shared across the four tasks. Stage~1 (video pretraining) is followed by Stage~2 (joint video+action fine-tuning from the Stage~1 checkpoint).}
  \label{tab:uva_maniskill_hparams_eval_settings}
  \resizebox{\textwidth}{!}{%
  \begin{tabular}{ll|cccc}
  \toprule
   &  & PickCube-v1 & PushCube-v1 & DrawTriangle-v1 & StackCube-v1 \\
  \midrule
  \multicolumn{6}{l}{\textbf{Hyperparameters (shared across stages unless noted)}} \\
   & optimizer                  & \multicolumn{4}{|c}{AdamW} \\
   & betas                      & \multicolumn{4}{|c}{(0.9, 0.95)} \\
   & weight\_decay              & \multicolumn{4}{|c}{0.02} \\
   & lr\_peak                   & \multicolumn{4}{|c}{1.00e-04} \\
   & lr\_scheduler              & \multicolumn{4}{|c}{cosine} \\
   & warmup\_steps              & \multicolumn{4}{|c}{1000} \\
   & batch\_size                & \multicolumn{4}{|c}{32} \\
   & mixed\_precision           & \multicolumn{4}{|c}{fp16} \\
   & EMA power / max            & \multicolumn{4}{|c}{0.75 / 0.9999} \\
   & action\_mask\_ratio        & \multicolumn{4}{|c}{0.5} \\
   & horizon                    & \multicolumn{4}{|c}{32} \\
   & num\_epochs (Stage~1 / Stage~2) & \multicolumn{4}{|c}{100 / 500} \\
   \midrule
  \multicolumn{6}{l}{\textbf{Evaluation Setting}} \\
   & n\_obs\_steps              & \multicolumn{4}{|c}{16} \\
   & model input frames         & \multicolumn{4}{|c}{4 (stride 4)} \\
   & n\_action\_steps           & \multicolumn{4}{|c}{8} \\
  \bottomrule
  \end{tabular}
  }
  \end{table}

\subsection{\texorpdfstring{$\pi_0$}{pi0}}
For $\pi_0$, we use the official openpi implementation\footnote{\url{https://github.com/Physical-Intelligence/openpi}} and fine-tune the $\pi_0$-base model with LoRA on each target task following the official $\pi_0$ LoRA schedule.
We start from the official LIBERO fine-tuning recipe, preserving the overall setup whenever possible, but fine-tune for 100k steps and evaluate checkpoints every 5k steps to obtain a reliable estimate of the baseline and reduce checkpoint-selection noise. DrawTriangle and StackCube were noticeably less stable than PickCube and PushCube in our setup, so we additionally tuned their optimization hyperparameters to improve training stability while keeping the rest of the pipeline unchanged.

% Table~\ref{tab:pi0_success_rate_by_step} reports the success rate of $\pi_0$ every 5k steps for the three ManiSkill tasks. The table clarifies how performance evolves with training and makes explicit which checkpoint range produces the strongest results for each task. 
%PushCube saturates early, PickCube benefits from mid-training checkpoints, and DrawTriangle remains the most unstable task, which motivates both the task-specific tuning and the longer 100k-step training horizon. 
Table~\ref{tab:pi0_hparams_eval_settings} summarizes the final hyperparameters and evaluation settings used to produce these results. PickCube and PushCube share one configuration, while DrawTriangle and StackCube share another. In particular, the DrawTriangle/StackCube pair uses a smaller batch size and a lower peak learning rate than the PickCube/PushCube pair, reflecting the additional tuning we introduced for stability.

% \begin{table}[h]
% \centering
% \caption{Success rate of $\pi_0$ evaluated every 5k training steps on the three ManiSkill tasks, from 5k to 100k steps. Bold values indicate the best checkpoint for each task.}
% \label{tab:pi0_success_rate_by_step}
% \resizebox{\textwidth}{!}{%
% \begin{tabular}{l|rrrrrrrrrrrrrrrrrrrr}
% \toprule
% Task & 5k & 10k & 15k & 20k & 25k & 30k & 35k & 40k & 45k & 50k & 55k & 60k & 65k & 70k & 75k & 80k & 85k & 90k & 95k & 100k \\
% \midrule
% PickCube     & 0.09 & 0.28 & 0.30 & 0.28 & 0.26 & 0.30 & \textbf{0.42} & 0.40 & 0.35 & 0.30 & 0.31 & 0.31 & 0.33 & 0.35 & 0.32 & 0.35 & 0.31 & 0.34 & 0.35 & 0.33 \\
% PushCube     & 0.93 & 0.99 & \textbf{1.00} & 1.00 & 0.98 & 0.98 & 0.98 & 0.98 & 0.97 & 0.99 & 0.98 & 0.99 & 0.97 & 0.99 & 0.98 & 0.99 & 0.98 & 0.98 & 0.99 & 0.98 \\
% DrawTriangle & 0.00 & 0.01 & 0.01 & 0.01 & 0.03 & 0.09 & 0.09 & \textbf{0.10} & 0.07 & 0.00 & 0.00 & 0.05 & 0.00 & 0.03 & 0.02 & 0.03 & 0.03 & 0.03 & 0.02 & 0.01 \\
% \bottomrule
% \end{tabular}%
% }
% \end{table}

\begin{table}[t]
\centering
\caption{Training hyperparameters and evaluation settings used for the $\pi_0$ baseline on PickCube-v1, PushCube-v1, DrawTriangle-v1, and StackCube-v1. All values are grouped by the PickCube/PushCube and DrawTriangle/StackCube task pairs.}
\label{tab:pi0_hparams_eval_settings}
\resizebox{\textwidth}{!}{%
\begin{tabular}{ll|cccc}
\toprule
 &  & PickCube-v1 & PushCube-v1 & DrawTriangle-v1 & StackCube-v1 \\
\midrule
\multicolumn{6}{l}{\textbf{Hyperparameters}} \\
 & batch\_size  & \multicolumn{2}{|c}{16} & \multicolumn{2}{c}{8} \\
 & lr\_peak     & \multicolumn{2}{|c}{1.00e-05} & \multicolumn{2}{c}{5.00e-06} \\
 & warmup\_steps & \multicolumn{2}{|c}{5000} & \multicolumn{2}{c}{1000} \\
 & decay\_steps & \multicolumn{2}{|c}{100k} & \multicolumn{2}{c}{100k} \\
 & decay\_lr    & \multicolumn{2}{|c}{0} & \multicolumn{2}{c}{0} \\
\midrule
\multicolumn{6}{l}{\textbf{Evaluation Setting}} \\
 & replan\_steps & \multicolumn{2}{|c}{5} & \multicolumn{2}{c}{5} \\
\bottomrule
\end{tabular}
}
\end{table}

\clearpage
\newpage

\section{Flow Matching Formulation}
\label{app:flow_matching_formulation}
Recently, diffusion models have shown promising results for modeling multi-modal distributions in generating images~\cite{SD1}, videos~\cite{sora}, and actions for robotic policies~\cite{DP}.
Flow matching~\cite{flowmatching} provides an alternative to diffusion models, offering improved training stability and faster inference while maintaining similar expressive power.
To retain these benefits while simplifying training and sampling, we leverage flow matching-based generative modeling to train both policy and dynamics model.

For clarity, we define our flow matching notation.
Let $x^{(\tau)}$ denote a flow sample at flow timestep $\tau \in [0,1]$, where the continuous path transforms from noise ($\tau=0$) to data ($\tau=1$).
As in the main text, trajectory subscripts are omitted for flow samples when the context is clear.

Let $q(x)$ be the target distribution, which is unknown.
For the sample $x_1 \sim q(\cdot)$, flow matching introduces the conditional probability density path $p_\tau(x^{(\tau)} | x_1)$ where $p_1 \approx q$ and $x^{(\tau)}$ is given by a time-dependent diffeomorphism, known as a continuous normalizing flow (CNF).
A common formulation is using a Gaussian conditional distribution:
\begin{equation}
    x^{(\tau)} = \tau x_1 + (1-\tau)x^{(0)}, \quad x^{(0)} \sim \mathcal{N}(0, I)
\end{equation}
Consider the conditional vector field $u^{(\tau)}(x^{(\tau)} | x_1)$ which indicates the change of the CNF $x^{(\tau)}$ with respect to flow timestep $\tau$.
\begin{equation}
    \frac{d}{d\tau} x^{(\tau)} = u^{(\tau)}(x^{(\tau)} | x_1)
\end{equation}
Conditional flow matching (CFM) proposes a supervised learning framework that trains a model estimate $\hat{u}^{(\tau)}(x^{(\tau)}, \tau)$ by regressing it onto the conditional vector field $u^{(\tau)}(x^{(\tau)} | x_1)$:
\begin{equation}
    \mathcal{L}_{\text{CFM}} = \mathbb{E}\left[\lVert u^{(\tau)}(x^{(\tau)} | x_1) - \hat{u}^{(\tau)}(x^{(\tau)}, \tau)\rVert^2\right]
\end{equation}
where $u^{(\tau)}(x^{(\tau)} | x_1)=x_1-x^{(0)}$ for the linear interpolation.
Although this target value has no explicit $\tau$ dependence for a linear path, the vector field remains indexed by $\tau$ because it is evaluated at the flow-time-dependent sample $x^{(\tau)}$.
The model does not observe $x_1$, so the minimizer of this objective is the marginal vector field $\mathbb{E}[u^{(\tau)}(x^{(\tau)} | x_1) \mid x^{(\tau)}]$, which transports the noise distribution to $q$.

\section{Theoretical Analysis}
\label{app:formal-theorem}

%-----------------------------------------------------------------------
\subsection{Preliminaries and Notation}
\label{app:prelim}
%-----------------------------------------------------------------------

We consider a behavior cloning setting where a robot observes
$o_t \in \mathcal{O}$ and executes action chunks
$a_t \in \mathcal{A} \subseteq \R^{d_a}$.
Two models are trained independently via flow matching:
\begin{itemize}[nosep]
    \item \textbf{Policy model} $\pi_\phi$: learns the conditional vector field
    for $p(a_t \mid o_t, o_{t+1})$.
    \item \textbf{Dynamics model} $f_\theta$: learns the conditional vector field
    for $p(o_{t+1} \mid o_t, a_t)$.
\end{itemize}
We denote the learned vector fields by $\uhata{\tau}, \uhatf{\tau}$, and the
ideal (perfectly trained) vector fields by $\uideala{\tau}, \uidealf{\tau}$.
The convention is: a hat $\hat{\cdot}$ marks learned-model quantities;
the superscript $*$ marks ideal (perfectly trained) quantities and, in $\ostarof{\tau}$, the dynamics-aligned fixed point of the learned map (Definition~\ref{def:aligned-pair}); for quantities
parametrized by flow time $\tau$, we write $^{*,(\tau)}$ with a comma separating
the ideal marker from the flow time.

Let $\tau \in [0,1]$ denote the flow time, with $\tau = 0$ corresponding to
pure noise and $\tau = 1$ to data.
Under linear flow matching, the interpolation path is
$x^{(\tau)} = (1-\tau)\,x^{(0)} + \tau\,x_1$
with conditional velocity $u^{*,(\tau)}(x^{(\tau)}, \tau) = x_1 - x^{(0)}$,
where $x^{(0)}$ denotes noise and $x_1$ denotes the data sample.
For the proofs, we adopt the discrete indexing $\tau_k := k\,d\tau = k/K$ with $d\tau=1/K$, and write $h:=d\tau$ for compactness.

Throughout this appendix, the current observation $o_t$ is fixed as a conditioning variable and suppressed from notation.
All norms $\norm{\cdot}$ are Euclidean unless stated otherwise.

\paragraph{Quantity Hierarchy.}
The analysis involves four distinct kinds of quantities; disambiguating them explicitly is essential for the end-to-end error analysis (Section~\ref{app:e2e}, Proposition~\ref{prop:gronwall}).

\begin{center}
\renewcommand{\arraystretch}{1.25}
\begin{tabular}{@{}>{\bfseries}l p{6.3cm} l@{}}
\toprule
Term & Meaning & Notation \\
\midrule
Expert    & Physically realized GT trajectory from $\mathcal{D}_{\text{expert}}$;
            no model, no estimate. & $a^{\expt}$, $o^{\expt}$ \\
Ideal     & Hypothetical perfect models with zero approximation error
            ($\varepsilon_\pi = \varepsilon_f = 0$); conditional OT vector fields. &
            $\uideala{\tau}, \uidealf{\tau}$, $\Epistarof{\tau}$, $\Fstarof{\tau}$ \\
Learned   & The actual trained models with bounded approximation error
            ($\varepsilon_\pi, \varepsilon_f > 0$). &
            $\uhata{\tau}, \uhatf{\tau}$, $\Epiof{\tau}$, $\Fof{\tau}$ \\
Reference & Hypothetical inference trajectory using \emph{learned} fields
            but with \emph{ideal} conditioning $\ostarof{\tau}$ at every step;
            an intermediate construct in the end-to-end bound (Section~\ref{app:e2e}). &
            $a^{\refsym}_K$, $a^{\refsym}_k$ \\
\bottomrule
\end{tabular}
\end{center}

The \textbf{reference trajectory} $\{a^{\refsym}_k\}$ is a theoretical device
used in the end-to-end deviation analysis of Section~\ref{app:e2e}: it denotes the action sequence produced by integrating the \emph{learned} policy field
$\uhata{\tau}$ but with the \emph{ideal} fixed-point conditioning
$\ostarof{\tau_k}$ at every step.
It is neither realizable at inference (the ideal fixed point is unknown) nor
the expert (which uses the ideal policy field). It exists solely to decompose
$\norm{\hat{a}^{\DAP} - a^{\expt}}$ into a tracking term (DAP vs.\ reference)
and a model-error term (reference vs.\ expert).

\paragraph{Flow Extrapolation.}
Given a flow sample $x^{(\tau)}$ at flow time $\tau$ and a learned vector field
$\hat{u}^{(\tau)}$, the flow extrapolation operator estimates the terminal sample:
\begin{equation}
    \Ext^{(\tau)}(x) \;:=\; x + (1-\tau) \cdot \hat{u}^{(\tau)}(x).
    \label{eq:extrap}
\end{equation}
Under the ideal field $u^{*,(\tau)}$, this recovers $x_1$ exactly.
With a learned field $\hat{u}^{(\tau)} = u^{*,(\tau)} + e^{(\tau)}$, the
extrapolation error is $(1-\tau)\norm{e^{(\tau)}}$.

\paragraph{Inference Variables.}
At flow step $k$ with discrete flow time $\tau_k = k/K$ and step size $h = 1/K$,
the algorithm maintains:

\begin{center}
\renewcommand{\arraystretch}{1.15}
\begin{tabular}{@{}ccl@{}}
\toprule
\textbf{Symbol} & \textbf{Space} & \textbf{Meaning} \\
\midrule
$\atauk{\tau_k}$ & $\mathcal{A}$ & Current action flow sample at time $\tau_k$ \\
$\otauk{\tau_k}$ & $\mathcal{O}$ & Current future-observation flow sample at time $\tau_k$ \\
$\ahattauk{\tau_k}$ & $\mathcal{A}$ & Value of $\hat{a}$ in Algorithm~\ref{alg:sample} after $k$ flow steps \\
$\ohattauk{\tau_k}$ & $\mathcal{O}$ & Value of $\hat{o}$ in Algorithm~\ref{alg:sample} after $k$ flow steps \\
\bottomrule
\end{tabular}
\end{center}

%-----------------------------------------------------------------------
\subsection{Definitions}
\label{app:definitions}
%-----------------------------------------------------------------------

\begin{defnA}[Policy and Dynamics Extrapolation Maps]
\label{def:extrap-maps}
At flow step $k$, the policy and dynamics extrapolation maps act on a
future-observation argument $\hat{o}$ and an action argument
$\hat{a}$ respectively:
\begin{alignat}{2}
    \Epiof{\tau}(\hat{o})
    &\;:=\; \atauk{\tau_k} + (1-\tau_k) \cdot
    \uhata{\tau_k}\!\bigl(\atauk{\tau_k}, \tau_k \mid \hat{o}\bigr),
    &\qquad &\text{(action extrapolation given future)}
    \label{eq:Epi} \\[2pt]
    \Efof{\tau}(\hat{a})
    &\;:=\; \otauk{\tau_k} + (1-\tau_k) \cdot
    \uhatf{\tau_k}\!\bigl(\otauk{\tau_k}, \tau_k \mid \hat{a}\bigr).
    &\qquad &\text{(future-obs extrapolation given action)}
    \label{eq:Ef}
\end{alignat}
Their \emph{ideal counterparts} $\Epistarof{\tau}$ and $\Efstarof{\tau}$ are
defined identically with $\uhata{\tau_k}, \uhatf{\tau_k}$ replaced by
$\uideala{\tau_k}, \uidealf{\tau_k}$.
\end{defnA}

\begin{defnA}[Dynamics-Alignment Map]
\label{def:alignment-map}
The \emph{dynamics-alignment map} is the composition of dynamics extrapolation
after policy extrapolation:
\begin{equation}
    \Fof{\tau}(\hat{o}) \;:=\; \Efof{\tau}\!\bigl(\Epiof{\tau}(\hat{o})\bigr).
    \label{eq:F}
\end{equation}
Starting from a future-observation candidate $\hat{o}$, the policy extrapolates
the action it would take given that future, and the dynamics extrapolates
the future that action would produce.
The output is the corrected future surrogate.
The \emph{ideal counterpart} is
$\Fstarof{\tau} := \Efstarof{\tau} \circ \Epistarof{\tau}$.
\end{defnA}

\begin{defnA}[Dynamics-Aligned Pair and Alignment Error]
\label{def:aligned-pair}
A pair $(\hat{a}^*, \hat{o}^*)$ is a \emph{dynamics-aligned pair} at flow time $\tau$
if $\hat{a}^* = \Epiof{\tau}(\hat{o}^*)$ and $\hat{o}^* = \Efof{\tau}(\hat{a}^*)$;
equivalently, $\hat{o}^*$ is a fixed point of $\Fof{\tau}$, denoted
$\ostarof{\tau}$.
At a dynamics-aligned pair, the policy's intended action and the dynamics' predicted
future are mutually consistent: the action targets the predicted future,
and the predicted future is the consequence of that action.

The \emph{alignment error} of a candidate $\hat{o}$ measures how far it deviates
from dynamics-alignment:
\begin{equation}
    R^{(\tau)}(\hat{o}) \;:=\; \norm{\hat{o} - \Fof{\tau}(\hat{o})}.
    \label{eq:alignment-error}
\end{equation}
\end{defnA}

%-----------------------------------------------------------------------
\subsection{Assumptions}
\label{app:assumptions}
%-----------------------------------------------------------------------

\begin{assA}[Cross-Conditioning Lipschitz Continuity]
\label{ass:cross-lip}
The learned vector fields are Lipschitz continuous in their
conditioning inputs: for all $(a, \tau, \hat{o}, \hat{o}')$ and
$(o, \tau, \hat{a}, \hat{a}')$,
\begin{align}
    \norm{\uhata{\tau}(a, \tau \mid \hat{o}) - \uhata{\tau}(a, \tau \mid \hat{o}')}
    &\leq L_\pi \norm{\hat{o} - \hat{o}'},
    \tag{A1} \label{eq:A1} \\
    \norm{\uhatf{\tau}(o, \tau \mid \hat{a}) - \uhatf{\tau}(o, \tau \mid \hat{a}')}
    &\leq L_f \norm{\hat{a} - \hat{a}'}.
    \tag{A2} \label{eq:A2}
\end{align}
\end{assA}

\begin{assA}[State Lipschitz Continuity]
\label{ass:state-lip}
The learned policy vector field is Lipschitz continuous in the state input:
\begin{equation}
    \norm{\uhata{\tau}(a, \tau \mid \hat{o}) - \uhata{\tau}(a', \tau \mid \hat{o})}
    \leq L_{\pi,a} \norm{a - a'}.
    \tag{A3} \label{eq:A3}
\end{equation}
\end{assA}

\begin{assA}[Regularity and Boundedness]
\label{ass:bounded}
The learned fields $\uhata{\tau}$ and $\uhatf{\tau}$ are $C^1$ on a compact
domain, with $\norm{\uhata{\tau}} \leq M$ and $\norm{\uhatf{\tau}} \leq M$
uniformly.
\end{assA}

\begin{assA}[Model Approximation]
\label{ass:approx}
The learned fields approximate the ideal conditional flow fields on the
algorithmic domain, i.e., at the flow samples and conditioning inputs at which
the analysis evaluates them, for $\tau \in [0, 1)$:
\begin{align}
    \sup\, \norm{\uhata{\tau}(a, \tau \mid \hat{o}) - \uideala{\tau}(a, \tau \mid \hat{o})}
    &\leq \varepsilon_\pi,
    \tag{A4} \label{eq:A4} \\
    \sup\,
    \norm{\uhatf{\tau}(o, \tau \mid \hat{a}) - \uidealf{\tau}(o, \tau \mid \hat{a})}
    &\leq \varepsilon_f,
    \tag{A5} \label{eq:A5}
\end{align}
with $\varepsilon_\pi$ and $\varepsilon_f$ independent of $K$.
The restriction $\tau \in [0,1)$ matches the algorithm's discrete schedule
$\tau_k = k/K \in [0, 1 - 1/K]$.
Because the ideal field (Lemma~\ref{lem:cancel}) scales as $1/(1-\tau)$ away
from its target, this condition, together with Assumption~\ref{ass:bounded},
requires the evaluated samples to lie within $\mathcal{O}(1-\tau)$ of their
targets. It is therefore a condition local to the sampling trajectories rather
than a uniform bound over a fixed compact set.
\end{assA}

\begin{assA}[Cross-Conditioning Contraction]
\label{ass:contraction}
The cross-conditioning Lipschitz constants satisfy
$L_\pi L_f < 1$.
\end{assA}

\begin{remA}
Assumptions~\ref{ass:cross-lip}--\ref{ass:bounded} are standard regularity
conditions for neural networks on compact domains.
Specifically, Assumptions~\ref{ass:cross-lip}--\ref{ass:state-lip} are taken globally on
$\R^{d_o}$ and $\R^{d_a}$ for the contraction argument; the compact-domain
hypothesis of Assumption~\ref{ass:bounded} is used only to obtain bounded-derivative constants.
The constants $L_\pi$ and $L_f$ measure the sensitivity of each vector field to its \emph{cross-conditioning} input---not the full input-output Lipschitz constant of the network---and are typically much smaller.
In practice, gradient norm clipping (e.g., clipping $\norm{\nabla_\theta \mathcal{L}}$ to $1.0$) prevents weight matrices from growing large during training, keeping these cross-conditioning sensitivities moderate.
The $C^1$ condition in Assumption~\ref{ass:bounded} ensures all partial derivatives (in $\tau$, state, and conditioning) exist and are bounded on the compact domain, which is needed for the fixed-point drift analysis (Proposition~\ref{prop:dap-convergence}).
Assumption~\ref{ass:contraction} requires that the feedback loop gain is less than one, which is natural for well-conditioned manipulation tasks.
This is empirically corroborated by Pan~et~al.~\cite{pan2025muchado}, who measure trained-policy Lipschitz constants on standard manipulation benchmarks via action-perturbation rollouts. 
On simpler, well-conditioned tasks (e.g., Push-T), flow policies measure below one across three architectures ($0.22$--$0.82$); on harder tasks the measured constants grow.
While their estimator differs from $L_\pi$ in detail---measuring the input-output Lipschitz of the converged policy rather than the cross-conditioning sensitivity of the conditional velocity field---the two are tightly related, and the small-Lipschitz regime empirically observed for flow policies on well-conditioned tasks provides direct support for $L_\pi L_f < 1$.
Even if Assumption~\ref{ass:contraction} is mildly violated, the results degrade gracefully: a pre-contraction phase reappears (see Remark~\ref{rem:phase1}), but the late-phase guarantees remain intact.
Section~\ref{app:measured} measures these constants for our models, and Assumption~\ref{ass:contraction} holds empirically on every simulation task.
\end{remA}

%-----------------------------------------------------------------------
\subsection{Main Theorem}
\label{app:main-theorem}
%-----------------------------------------------------------------------

\begin{thmA}[Dynamics-Aligned Contraction]
\label{thm:formal}
Under Assumptions~\ref{ass:cross-lip}--\ref{ass:contraction},
let $K$ denote the number of flow steps and $h = 1/K$ the step size.
Let $(a^{\expt}, o^{\expt})$ denote a physically realized expert
demonstration pair.

\medskip\noindent
\textbf{(i) Ideal foundation.}\;
In the deterministic setting (Remark~\ref{rem:scope}) and under ideal models $(\uideala{\tau}, \uidealf{\tau})$, the dynamics-alignment
map reduces to the $\tau$-independent composition of forward dynamics $d$ and
expert behavioral map $g$:
\begin{equation}
    \Fstarof{\tau}(\hat{o}) \;=\; d\bigl(o_t,\, g(o_t, \hat{o})\bigr).
    \label{eq:ideal-F}
\end{equation}
Every expert pair $(a^{\expt}, o^{\expt})$ is a dynamics-aligned pair:
$o^{\expt} = \Fstarof{\tau}(o^{\expt})$ for every $\tau \in [0,1)$.

\medskip\noindent
\textbf{(ii) Learned contraction and expert proximity.}\;
Under learned models, $\Fof{\tau}$ is a contraction with constant
$\kappa^{(\tau)} = (1-\tau)^2 L_\pi L_f < 1$ for all $\tau \in [0,1]$.
For every $\tau \in [0,1)$, the unique fixed point $\ostarof{\tau}$ satisfies:
\begin{equation}
    \norm{\ostarof{\tau} - o^{\expt}} \;\leq\;
    \frac{(1-\tau)\bigl(\varepsilon_f + (1-\tau)L_f\,\varepsilon_\pi\bigr)}
    {1 - (1-\tau)^2 L_\pi L_f}
    \;=:\; \Delta_\tau.
    \label{eq:bridge}
\end{equation}

\medskip\noindent
\textbf{(iii) DAP convergence (Gauss-Seidel idealization).}\;
Under the Gauss-Seidel idealization where each step applies
$\ohattauk{\tau_{k+1}} = \Fof{\tau_k}(\ohattauk{\tau_k})$
(Proposition~\ref{prop:jacobi} extends this to the Jacobi update of Algorithm~\ref{alg:sample}),
the conditioning mismatch
$\delta_k := \norm{\ohattauk{\tau_k} - \ostarof{\tau_k}}$
satisfies:
\begin{equation}
    \delta_{k+1} \leq \kappa^{(\tau_k)}\,\delta_k + C_{\mathrm{drift}}\, h,
    \label{eq:dap-recurrence}
\end{equation}
where $C_{\mathrm{drift}}$ is a finite constant determined by the $C^1$ bounds
of $\uhata{\tau}, \uhatf{\tau}$ on the compact domain
(Assumption~\ref{ass:bounded}).
The cumulative conditioning mismatch is bounded by:
\begin{equation}
    \sum_{k=0}^{K-1} h\,\delta_k \;\leq\; \eta_K
    \;:=\; \frac{\delta_0 + C_{\mathrm{drift}}}{K(1 - L_\pi L_f)},
    \label{eq:etaK}
\end{equation}
where $\delta_0 \leq D_\Omega$ ($D_\Omega$ being the diameter of
the observation domain).
In particular, $\eta_K = \mathcal{O}(1/K) \to 0$ as $K \to \infty$.

\medskip\noindent
\textbf{(iv) End-to-end bound.}\;
The final action errors are:
\begin{align}
    \norm{\hat{a}^{\DAP} - a^{\expt}}
    &\;\leq\; \Gamma\,\eta_K \;+\; \Gamma'\,\mathcal{M},
    \label{eq:e2e-dap} \\[4pt]
    \norm{\hat{a}^{\FC} - a^{\expt}}
    &\;\leq\; \Gamma\,C_{\FC} \;+\; \Gamma'\,\mathcal{M},
    \quad C_{\FC} := \textstyle\sum_{k=0}^{K-1}h\norm{\hat{o}^{\FC} - \ostarof{\tau_k}(\atauk{\tau_k})}
    \;\geq\; \delta_{\FC},
    \label{eq:e2e-fc}
\end{align}
where:
\begin{itemize}[nosep,leftmargin=2em]
    \item $\Gamma = e^{L_{\pi,a} + L_\pi \bar{L}_{o^*}} \cdot L_\pi$,
    with $\bar{L}_{o^*}$ the Lipschitz constant of the fixed-point map
    $a \mapsto \ostarof{\tau}(a)$, and $\Gamma' = e^{L_{\pi,a}}$;
    \item $\displaystyle\mathcal{M} = \varepsilon_\pi
    + \frac{L_\pi(\varepsilon_f + L_f\,\varepsilon_\pi)}
    {2(1 - L_\pi L_f)} \cdot \frac{K+1}{K}$
    is the cumulative model error;
    \item $\delta_{\FC} = \inf_{\tau, a} \norm{\hat{o}^{\FC} - \ostarof{\tau}(a)} > 0$
    is the conditioning mismatch of the fixed surrogate $\hat{o}^{\FC}$,
    independent of $K$.
\end{itemize}
As $K \to \infty$: the DAP tracking term $\Gamma\,\eta_K \to 0$,
so the bound converges to pure model error $\Gamma'\mathcal{M}$.
The fixed conditioning tracking term $\Gamma\,C_{\FC}$ does not vanish,
as $C_{\FC} \geq \delta_{\FC} > 0$;
DAP therefore admits a strictly tighter guarantee for sufficiently large $K$.
\end{thmA}

\begin{proof}
The proof assembles results from Sections~\ref{app:ideal}--\ref{app:e2e}.
Part~(i) follows from Lemma~\ref{lem:cancel} and Lemma~\ref{lem:expert-fp}.
Part~(ii) combines Proposition~\ref{prop:contraction} and
Proposition~\ref{prop:bridge}.
Part~(iii) is Proposition~\ref{prop:dap-convergence}.
Part~(iv) is Corollary~\ref{cor:e2e}.
\end{proof}

\begin{remA}[Interpretation]
The $(1-\tau)^2$ factor in $\kappa^{(\tau)}$ is not an assumption---it is an
algebraic consequence of flow extrapolation (Proposition~\ref{prop:contraction}).
As $\tau \to 1$, the flow sample already encodes most information about the data,
making conditioning progressively less influential.
DAP's contraction strengthens automatically in the late phase.
An intentionally bad initial conditioning enters the bound as $\delta_0$
in~\eqref{eq:dap-recurrence}, multiplied by the contraction product
$\prod\kappa^{(\tau_j)} \leq (L_\pi L_f)^k$---geometrically decaying.
Bad initialization is a transient, not a persistent loop.
\end{remA}

\begin{remA}[Fixed conditioning as open-loop inference]
\label{rem:FC-open-loop}
Fixed conditioning produces a future surrogate $\hat{o}^{\FC}$ once---typically
via a video prediction model or a single forward dynamics pass---and holds it
constant.
The mismatch $\delta_{\FC}$ arises because $\hat{o}^{\FC}$ is generated
without knowledge of the policy's action, so it cannot satisfy the
dynamics-alignment condition $\hat{o} = \Fof{\tau}(\hat{o})$.
This encompasses all video-conditioned policies and any scheme where future
prediction is decoupled from action generation.
FC never applies $\Fof{\tau}$ and therefore has no contraction mechanism;
its bound cannot be tightened by increasing $K$.
\end{remA}

\begin{remA}
\label{rem:phase1}
If Assumption~\ref{ass:contraction} is relaxed to $L_\pi L_f \geq 1$,
contraction holds only for
$\tau > \tau^* := 1 - 1/\sqrt{L_\pi L_f} \in (0,1)$.
A pre-contraction phase ($\tau \leq \tau^*$) reappears where neither method
has a structural advantage, adding a symmetric error term to both bounds.
The late-phase conclusions---DAP's convergence and fixed conditioning's
persistence---remain unchanged.
\end{remA}

%-----------------------------------------------------------------------
\subsection{Ideal Foundation}
\label{app:ideal}
%-----------------------------------------------------------------------

\begin{lemA}[Ideal Extrapolation Cancellation]
\label{lem:cancel}
In the deterministic setting, the ideal conditional vector fields
(conditional optimal transport fields) satisfy, for all $a, \hat{a} \in \R^{d_a}$,
$o, \hat{o} \in \R^{d_o}$, and $\tau \in [0,1)$:
\begin{equation}
    \uideala{\tau}(a, \tau \mid \hat{o})
    = \frac{g(o_t, \hat{o}) - a}{1-\tau}, \qquad
    \uidealf{\tau}(o, \tau \mid \hat{a})
    = \frac{d(o_t, \hat{a}) - o}{1-\tau}.
    \label{eq:ideal-v}
\end{equation}
These are defined for \emph{all} states $(a, o)$, not only on the
interpolation path---this is standard in the conditional OT framework.
Consequently, for any flow samples $(\atauk{\tau_k}, \otauk{\tau_k})$ and
every $\tau_k \in [0,1)$:
\begin{equation}
    \Epistarof{\tau_k}(\hat{o}) = g(o_t, \hat{o}), \qquad
    \Efstarof{\tau_k}(\hat{a}) = d(o_t, \hat{a}).
    \label{eq:cancel}
\end{equation}
\end{lemA}

\begin{proof}
The conditional OT vector field transporting any point $a$ toward
the target $g(o_t, \hat{o})$ in remaining time $(1-\tau)$ is
$(g(o_t, \hat{o}) - a)/(1-\tau)$---this holds for all $a$,
not only on the interpolation path.
Substituting into the extrapolation operator at the algorithm's
current flow sample $\atauk{\tau_k}$:
\[
    \Epistarof{\tau_k}(\hat{o})
    = \atauk{\tau_k} + (1-\tau_k) \cdot \frac{g(o_t, \hat{o}) - \atauk{\tau_k}}{1-\tau_k}
    = g(o_t, \hat{o}).
\]
The $(1-\tau_k)$ factors cancel exactly, making the result independent of
$\atauk{\tau_k}$ and $\tau_k$.
The argument for $\Efstarof{\tau_k}$ is identical.
\end{proof}

\begin{lemA}[Expert Pair as Ideal Dynamics-Aligned Pair]
\label{lem:expert-fp}
The ideal dynamics-alignment map is $\tau$-independent for every $\tau \in [0,1)$:
\begin{equation}
    \Fstarof{\tau}(\hat{o}) = d\bigl(o_t, g(o_t, \hat{o})\bigr).
    \label{eq:Fstar}
\end{equation}
Every expert demonstration pair $(a^{\expt}, o^{\expt})$ satisfies
$o^{\expt} = \Fstarof{\tau}(o^{\expt})$
and $a^{\expt} = \Epistarof{\tau}(o^{\expt})$
for every $\tau \in [0,1)$.
\end{lemA}

\begin{proof}
Equation~\eqref{eq:Fstar} follows from substituting~\eqref{eq:cancel}
into $\Fstarof{\tau} = \Efstarof{\tau} \circ \Epistarof{\tau}$.
Since $(a^{\expt}, o^{\expt})$
is a physically realized trajectory:
\begin{equation}
    o^{\expt} = d(o_t, a^{\expt})
    \quad\text{and}\quad
    a^{\expt} = g(o_t, o^{\expt}).
    \label{eq:physical}
\end{equation}
Substituting the second into the first:
$o^{\expt} = d(o_t, g(o_t, o^{\expt})) = \Fstarof{\tau}(o^{\expt})$.
This is a tautological consequence of the trajectory
being physically realized, not an assumption.
\end{proof}

\begin{remA}[Scope]
\label{rem:scope}
Lemmas~\ref{lem:cancel} and~\ref{lem:expert-fp}, and the expert-proximity bounds
that build on them (Proposition~\ref{prop:bridge} and the terms of
Corollary~\ref{cor:e2e} that use it), assume deterministic conditionals:
a single expert action for each $(o_t, o_{t+1})$ and a single future for each
$(o_t, a_t)$.
When several valid actions exist for the same $(o_t, o_{t+1})$, the ideal
extrapolation returns their conditional mean given the current flow sample
rather than a single endpoint, so the cancellation in Lemma~\ref{lem:cancel}
does not hold in general; we do not extend these bounds to that case.
The contraction and tracking results for the learned maps
(Propositions~\ref{prop:contraction}, \ref{prop:fp-existence},
and~\ref{prop:dap-convergence}--\ref{prop:gronwall}) do not rely on this assumption.
\end{remA}

\begin{remA}[Uniqueness under additional regularity]
The Jacobian of $\Fstarof{\tau}$ is $\nabla_{\hat{o}} \Fstarof{\tau} = J_d \cdot J_g$,
where $J_d = \nabla_{\hat{a}} d(o_t, \cdot)$ and $J_g = \nabla_{\hat{o}} g(o_t, \cdot)$.
If the task-intrinsic loop gain $\kappa^* := \norm{J_d}_\op \norm{J_g}_\op < 1$,
Banach's theorem guarantees global uniqueness.
Parts~(ii)--(iv) of the main theorem do not require $\kappa^* < 1$:
the learned map's contraction arises from the $(1-\tau)^2$ factor.
\end{remA}

%-----------------------------------------------------------------------
\subsection{Learned Contraction and Expert Proximity}
\label{app:contraction}
%-----------------------------------------------------------------------

\begin{propA}[Automatic Contraction of the Learned Map]
\label{prop:contraction}
Under Assumptions~\ref{ass:cross-lip} and~\ref{ass:contraction},
$\Fof{\tau}$ is a contraction with constant
$\kappa^{(\tau)} = (1-\tau)^2 L_\pi L_f < 1$ for all $\tau \in [0,1]$:
\begin{equation}
    \norm{\Fof{\tau}(\hat{o}) - \Fof{\tau}(\hat{o}')}
    \;\leq\; \kappa^{(\tau)} \norm{\hat{o} - \hat{o}'}
    \quad \text{for all } \hat{o}, \hat{o}' \in \R^{d_o}.
    \label{eq:kappa}
\end{equation}
\end{propA}

\begin{proof}
We use Lipschitz composition directly (no differentiability required).
First, $\Epiof{\tau}$ is $(1-\tau)L_\pi$-Lipschitz in its conditioning argument:
for any $\hat{o}, \hat{o}'$,
\[
    \norm{\Epiof{\tau}(\hat{o}) - \Epiof{\tau}(\hat{o}')}
    = (1-\tau)\norm{\uhata{\tau}(\atauk{\tau_k}, \tau_k \mid \hat{o}) - \uhata{\tau}(\atauk{\tau_k}, \tau_k \mid \hat{o}')}
    \leq (1-\tau)L_\pi\norm{\hat{o} - \hat{o}'}.
\]
Similarly, $\Efof{\tau}$ is $(1-\tau)L_f$-Lipschitz in its argument.
Composing:
\begin{align*}
    \norm{\Fof{\tau}(\hat{o}) - \Fof{\tau}(\hat{o}')}
    &= \norm{\Efof{\tau}(\Epiof{\tau}(\hat{o})) - \Efof{\tau}(\Epiof{\tau}(\hat{o}'))} \\
    &\leq (1-\tau)L_f \cdot \norm{\Epiof{\tau}(\hat{o}) - \Epiof{\tau}(\hat{o}')} \\
    &\leq (1-\tau)L_f \cdot (1-\tau)L_\pi \cdot \norm{\hat{o} - \hat{o}'}
    = \kappa^{(\tau)}\norm{\hat{o} - \hat{o}'}.
    \qedhere
\end{align*}
\end{proof}

\begin{remA}[Contrast with the ideal case]
Under ideal models, the $(1-\tau)$ factors in the extrapolation Lipschitz constants
cancel with the $1/(1-\tau)$ poles in the ideal vector field~\eqref{eq:ideal-v},
yielding the $\tau$-independent contraction constant
$\kappa^* = \norm{J_d}_\op\norm{J_g}_\op$.
Under learned models, the bounded vector fields
(Assumption~\ref{ass:bounded}) prevent this cancellation,
so the $(1-\tau)^2$ factor survives---advantageously,
as the learned map contracts more strongly for $\tau$ close to $1$.
\end{remA}

\begin{propA}[Existence of Dynamics-Aligned Fixed Point]
\label{prop:fp-existence}
Under Assumptions~\ref{ass:cross-lip} and~\ref{ass:contraction},
for every $\tau \in [0,1]$, $\Fof{\tau}$ admits a unique fixed point
$\ostarof{\tau}$.
For any $\hat{o} \in \R^{d_o}$:
\begin{equation}
    \norm{\hat{o} - \ostarof{\tau}}
    \;\leq\; \frac{R^{(\tau)}(\hat{o})}{1 - \kappa^{(\tau)}}.
    \label{eq:fp-dist}
\end{equation}
\end{propA}

\begin{proof}
$\Fof{\tau}$ is a contraction on the complete metric space
$(\R^{d_o}, \norm{\cdot})$ by Proposition~\ref{prop:contraction}.
Banach's fixed-point theorem yields existence and uniqueness.
For~\eqref{eq:fp-dist}:
\[
    \norm{\hat{o} - \ostarof{\tau}}
    \leq \norm{\hat{o} - \Fof{\tau}(\hat{o})} + \norm{\Fof{\tau}(\hat{o}) - \Fof{\tau}(\ostarof{\tau})}
    \leq R^{(\tau)}(\hat{o}) + \kappa^{(\tau)} \norm{\hat{o} - \ostarof{\tau}}.
\]
Rearranging: $(1-\kappa^{(\tau)})\norm{\hat{o} - \ostarof{\tau}} \leq R^{(\tau)}(\hat{o})$.
\end{proof}

\begin{propA}[Model Error Bridge]
\label{prop:bridge}
Under Assumptions~\ref{ass:cross-lip},~\ref{ass:approx},
and~\ref{ass:contraction}, for every $\tau \in [0,1)$,
the learned fixed point $\ostarof{\tau}$
and the expert observation $o^{\expt}$ satisfy:
\begin{equation}
    \norm{\ostarof{\tau} - o^{\expt}}
    \;\leq\;
    \frac{(1-\tau)\bigl(\varepsilon_f + (1-\tau)L_f\,\varepsilon_\pi\bigr)}
    {1 - (1-\tau)^2 L_\pi L_f}
    \;=:\; \Delta_\tau.
    \label{eq:bridge-bound}
\end{equation}
In particular, $\Delta_\tau = O(1-\tau) \to 0$ as $\tau \to 1$.
\end{propA}

\begin{proof}
Since $\ostarof{\tau} = \Fof{\tau}(\ostarof{\tau})$ and
$o^{\expt} = \Fstarof{\tau}(o^{\expt})$ (Lemma~\ref{lem:expert-fp}):
\begin{equation}
    \norm{\ostarof{\tau} - o^{\expt}}
    \leq \underbrace{\norm{\Fof{\tau}(\ostarof{\tau})
    - \Fof{\tau}(o^{\expt})}}_{\text{(I)}}
    + \underbrace{\norm{\Fof{\tau}(o^{\expt})
    - \Fstarof{\tau}(o^{\expt})}}_{\text{(II)}}.
    \label{eq:bridge-triangle}
\end{equation}

\textbf{Term~(I).}\;
By contraction: $\text{(I)} \leq \kappa^{(\tau)} \norm{\ostarof{\tau} - o^{\expt}}$.

\textbf{Term~(II).}\;
For any $\hat{o}$, let $\hat{a} := \Epiof{\tau}(\hat{o})$ and
$\hat{a}^* := \Epistarof{\tau}(\hat{o})$.
Decomposing $\Fof{\tau}(\hat{o}) - \Fstarof{\tau}(\hat{o})
= \Efof{\tau}(\hat{a}) - \Efstarof{\tau}(\hat{a}^*)$:
\begin{align}
    \norm{\Efof{\tau}(\hat{a}) - \Efof{\tau}(\hat{a}^*)}
    &\leq (1-\tau)L_f \cdot (1-\tau)\varepsilon_\pi
    = (1-\tau)^2 L_f\,\varepsilon_\pi,
    \label{eq:term-II-a} \\
    \norm{\Efof{\tau}(\hat{a}^*) - \Efstarof{\tau}(\hat{a}^*)}
    &= (1-\tau)\norm{\uhatf{\tau} - \uidealf{\tau}}
    \leq (1-\tau)\varepsilon_f.
    \label{eq:term-II-b}
\end{align}
The first line uses the $(1-\tau)L_f$-Lipschitz property of $\Efof{\tau}$
(established in the proof of Proposition~\ref{prop:contraction},
via~\eqref{eq:A2}) together with
$\norm{\hat{a} - \hat{a}^*} \leq (1-\tau)\varepsilon_\pi$ (from~\eqref{eq:A4}).
Combining: $\text{(II)} \leq (1-\tau)(\varepsilon_f + (1-\tau)L_f\,\varepsilon_\pi)$.

Substituting into~\eqref{eq:bridge-triangle} and rearranging:
$(1 - \kappa^{(\tau)})\norm{\ostarof{\tau} - o^{\expt}}
\leq (1-\tau)(\varepsilon_f + (1-\tau)L_f\,\varepsilon_\pi)$.
\end{proof}

%-----------------------------------------------------------------------
\subsection{DAP Convergence}
\label{app:convergence}
%-----------------------------------------------------------------------

\textbf{DAP inference} (Jacobi-style update):
at each flow step $k$, both extrapolated estimates are updated simultaneously:
$\ahattauk{\tau_{k+1}} = \Epiof{\tau_k}(\ohattauk{\tau_k})$,\;
$\ohattauk{\tau_{k+1}} = \Efof{\tau_k}(\ahattauk{\tau_k})$.

\begin{remA}[Jacobi vs.\ Gauss-Seidel]
\label{rem:jacobi}
Our analysis uses the Gauss-Seidel variant
$\ohattauk{\tau_{k+1}} = \Efof{\tau_k}(\Epiof{\tau_k}(\ohattauk{\tau_k})) = \Fof{\tau_k}(\ohattauk{\tau_k})$,
where each step is one application of the dynamics-alignment map.
The per-step discrepancy between Jacobi and Gauss-Seidel is:
\[
    \norm{\Efof{\tau_k}(\ahattauk{\tau_k}) - \Efof{\tau_k}(\ahattauk{\tau_{k+1}})}
    \leq (1-\tau_k) L_f \norm{\ahattauk{\tau_k} - \ahattauk{\tau_{k+1}}}.
\]
The factor $(1-\tau_k)$ suppresses this discrepancy in the late phase
where contraction matters most.
Proposition~\ref{prop:dap-convergence} is stated for the Gauss-Seidel variant,
and Proposition~\ref{prop:jacobi} extends it to the Jacobi update.
\end{remA}

\begin{propA}[DAP Alignment Convergence]
\label{prop:dap-convergence}
Under Assumptions~\ref{ass:cross-lip}--\ref{ass:contraction},
define $\delta_k := \norm{\ohattauk{\tau_k} - \ostarof{\tau_k}}$.

\medskip\noindent
\textbf{(a) Contraction recurrence.}
\begin{equation}
    \delta_{k+1} \leq \kappa^{(\tau_k)}\,\delta_k + C_{\mathrm{drift}} h,
    \label{eq:recurrence}
\end{equation}
where
$C_{\mathrm{drift}}$ is a finite constant determined by the $C^1$ derivative bounds
of $\uhata{\tau}, \uhatf{\tau}$ and the contraction gap $1 - L_\pi L_f$
(see proof below for derivation).

\medskip\noindent
\textbf{(b) Explicit bound.}
\begin{equation}
    \delta_k \;\leq\;
    \Biggl(\prod_{j=0}^{k-1} \kappa^{(\tau_j)}\Biggr) \delta_0
    \;+\; \frac{C_{\mathrm{drift}} h}{1 - L_\pi L_f}.
    \label{eq:delta-explicit}
\end{equation}

\medskip\noindent
\textbf{(c) Cumulative conditioning mismatch.}
\begin{equation}
    \sum_{k=0}^{K-1} h\,\delta_k
    \;\leq\; \eta_K
    \;:=\; \frac{\delta_0 + C_{\mathrm{drift}}}{K(1 - L_\pi L_f)},
    \label{eq:eta-explicit}
\end{equation}
where $\delta_0 \leq D_\Omega$ ($D_\Omega$ being the diameter of the compact domain).
In particular, $\eta_K = \mathcal{O}(1/K) \to 0$ as $K \to \infty$.

\medskip\noindent
\textbf{(d) Decay of the contraction product.}
For all $k$, the product decays at least geometrically:
$\prod_{j=0}^{k-1}\kappa^{(\tau_j)} \leq (L_\pi L_f)^k$.
At the terminal step $k = K$, a sharper bound holds:
\begin{equation}
    \prod_{j=0}^{K-1} \kappa^{(\tau_j)}
    = (L_\pi L_f)^K \Bigl(\frac{K!}{K^K}\Bigr)^{\!2}.
    \label{eq:super-exp}
\end{equation}
By Stirling's approximation, $(K!/K^K)^2 \approx 2\pi K \cdot e^{-2K}$.
Since $L_\pi L_f < 1$, the terminal product decays
faster than $e^{-2K}$.
\end{propA}

\begin{proof}
\textbf{Part~(a).}\;
In the Gauss-Seidel idealization, $\ohattauk{\tau_{k+1}} = \Fof{\tau_k}(\ohattauk{\tau_k})$:
\begin{align*}
    \delta_{k+1}
    &= \norm{\Fof{\tau_k}(\ohattauk{\tau_k}) - \ostarof{\tau_{k+1}}} \\
    &\leq \underbrace{\norm{\Fof{\tau_k}(\ohattauk{\tau_k}) - \Fof{\tau_k}(\ostarof{\tau_k})}}_{\leq\, \kappa^{(\tau_k)}\,\delta_k}
    + \underbrace{\norm{\ostarof{\tau_k} - \ostarof{\tau_{k+1}}}}_{\leq\, C_{\mathrm{drift}} h}.
\end{align*}
The first term follows from contraction (Proposition~\ref{prop:contraction}).
For the second term: between consecutive steps,
$\norm{\atauk{\tau_{k+1}} - \atauk{\tau_k}} \leq hM$,\;
$\norm{\otauk{\tau_{k+1}} - \otauk{\tau_k}} \leq hM$,\;
and $|\tau_{k+1} - \tau_k| = h$.
Since $\kappa^{(\tau_k)} < 1$ ensures $I - \nabla_{\hat{o}} \Fof{\tau_k}$ is invertible
(the Jacobian $\nabla_{\hat{o}} \Fof{\tau}$ exists by the $C^1$ condition
in Assumption~\ref{ass:bounded}),
the implicit function theorem gives
$\norm{\ostarof{\tau_k} - \ostarof{\tau_{k+1}}} \leq C_{\mathrm{drift}}\, h$.
Here $C_{\mathrm{drift}} = C_{\mathrm{map}} / (1 - L_\pi L_f)$,
where $C_{\mathrm{map}}$ is the uniform bound on the sensitivity
of $\Fof{\tau}(\ostarof{\tau})$ to changes in
$(\atauk{\tau_k}, \otauk{\tau_k}, \tau_k)$---finite
by the $C^1$ condition on a compact domain (Assumption~\ref{ass:bounded})---and
$(1 - L_\pi L_f)^{-1}$ comes from inverting
$I - \nabla_{\hat{o}} \Fof{\tau}$ via the Neumann series.

\medskip
\textbf{Part~(b).}\;
Iterating~\eqref{eq:recurrence} from $k = 0$ with $\kappa^{(\tau_j)} \leq L_\pi L_f < 1$:
\[
    \delta_k
    \leq \Biggl(\prod_{j=0}^{k-1}\kappa^{(\tau_j)}\Biggr)\delta_0
    + C_{\mathrm{drift}} h \sum_{i=0}^{k-1} \prod_{j=i+1}^{k-1}\kappa^{(\tau_j)}
    \leq \Biggl(\prod_{j=0}^{k-1}\kappa^{(\tau_j)}\Biggr)\delta_0
    + \frac{C_{\mathrm{drift}} h}{1 - L_\pi L_f}.
\]

\medskip
\textbf{Part~(c).}\;
Using $\prod_{j=0}^{k-1}\kappa^{(\tau_j)} \leq (L_\pi L_f)^k$:
\[
    \sum_{k=0}^{K-1} h\,\delta_k
    \leq \frac{\delta_0}{K} \sum_{k=0}^{K-1}(L_\pi L_f)^k
    + \frac{C_{\mathrm{drift}}}{K(1-L_\pi L_f)}
    \leq \frac{\delta_0}{K(1-L_\pi L_f)}
    + \frac{C_{\mathrm{drift}}}{K(1-L_\pi L_f)}.
\]

\medskip
\textbf{Part~(d).}\;
With $\tau_j = j/K$:
$\prod_{j=0}^{K-1}\kappa^{(\tau_j)}
= (L_\pi L_f)^K \prod_{j=0}^{K-1}((K-j)/K)^{2}
= (L_\pi L_f)^K (K!/K^K)^{2}$.
By Stirling, $K! \approx \sqrt{2\pi K}\,(K/e)^K$,
giving $(K!/K^K)^{2} \approx 2\pi K \cdot e^{-2K}$.
\end{proof}

\begin{propA}[Jacobi Update]
\label{prop:jacobi}
Under Assumptions~\ref{ass:cross-lip}--\ref{ass:contraction}, taking $L_\pi, L_f > 0$ without loss of generality, let
$\alpha := \sqrt{L_\pi/L_f}$ and $\norm{(a, o)}_\alpha := \max\bigl(\norm{a}/\alpha,\, \norm{o}\bigr)$.
At every flow time $\tau$, the Jacobi map applied by Algorithm~\ref{alg:sample},
\begin{equation}
    J^{(\tau)}(\hat{a}, \hat{o}) \;:=\; \bigl(\Epiof{\tau}(\hat{o}),\, \Efof{\tau}(\hat{a})\bigr),
    \label{eq:jacobi-map}
\end{equation}
is a contraction in $\norm{\cdot}_\alpha$ with constant $(1-\tau)\sqrt{L_\pi L_f}$,
and its unique fixed point is the dynamics-aligned pair of Definition~\ref{def:aligned-pair}.
Hence Proposition~\ref{prop:dap-convergence}(a)--(c) and Corollary~\ref{cor:e2e} hold for
Algorithm~\ref{alg:sample}, with the conditioning mismatch measured in $\norm{\cdot}_\alpha$,
$\kappa^{(\tau)}$ replaced by $(1-\tau)\sqrt{L_\pi L_f}$, and $L_\pi L_f$ replaced by $\sqrt{L_\pi L_f}$.
\end{propA}

\begin{proof}
By the proof of Proposition~\ref{prop:contraction}, $\Epiof{\tau}$ is $(1-\tau)L_\pi$-Lipschitz
and $\Efof{\tau}$ is $(1-\tau)L_f$-Lipschitz in their arguments. For any two pairs,
\begin{align*}
    \norm{J^{(\tau)}(\hat{a}, \hat{o}) - J^{(\tau)}(\hat{a}', \hat{o}')}_\alpha
    &= \max\Bigl(\tfrac{1}{\alpha}\norm{\Epiof{\tau}(\hat{o}) - \Epiof{\tau}(\hat{o}')},\,
       \norm{\Efof{\tau}(\hat{a}) - \Efof{\tau}(\hat{a}')}\Bigr) \\
    &\leq (1-\tau)\max\Bigl(\tfrac{L_\pi}{\alpha}\norm{\hat{o} - \hat{o}'},\,
       L_f\alpha\,\tfrac{1}{\alpha}\norm{\hat{a} - \hat{a}'}\Bigr) \\
    &\leq (1-\tau)\sqrt{L_\pi L_f}\,\norm{(\hat{a} - \hat{a}', \hat{o} - \hat{o}')}_\alpha,
\end{align*}
since $L_\pi/\alpha = L_f\alpha = \sqrt{L_\pi L_f}$.
A fixed point satisfies $\hat{a}^* = \Epiof{\tau}(\hat{o}^*)$ and $\hat{o}^* = \Efof{\tau}(\hat{a}^*)$,
which is Definition~\ref{def:aligned-pair}.
The proof of Proposition~\ref{prop:dap-convergence} then carries over with the conditioning mismatch
measured in $\norm{\cdot}_\alpha$, which upper-bounds $\norm{\ohattauk{\tau_k} - \ostarof{\tau_k}}$,
and with the Neumann-series inversion of $I - \nabla J^{(\tau)}$ using the new contraction constant.
For the terminal estimate in Corollary~\ref{cor:e2e}, the Jacobi update gives
$\ohattauk{\tau_K} = \Efof{\tau_{K-1}}(\ahattauk{\tau_{K-1}})$ and $\ostarof{\tau_{K-1}} = \Efof{\tau_{K-1}}(\hat{a}^*)$,
so their distance is at most $(1-\tau_{K-1})L_f\alpha = (1-\tau_{K-1})\sqrt{L_\pi L_f}$ times the conditioning mismatch in $\norm{\cdot}_\alpha$.
\end{proof}

%-----------------------------------------------------------------------
\subsection{Fixed Conditioning}
\label{app:frozen}
%-----------------------------------------------------------------------

\textbf{Fixed conditioning inference:}\;
a surrogate $\hat{o}^{\FC}$ is produced once and held constant:
$\atauk{\tau_{k+1}} = \atauk{\tau_k} + h \cdot \uhata{\tau_k}(\atauk{\tau_k}, \tau_k \mid \hat{o}^{\FC})$.

\begin{propA}[Fixed Conditioning Bound Does Not Vanish]
\label{prop:fc-persist}
For fixed conditioning with surrogate $\hat{o}^{\FC}$,
let $\delta_{\FC} := \inf_{\tau, a} \norm{\hat{o}^{\FC} - \ostarof{\tau}(a)}$,
where the infimum is over all flow times and action states.
If $\delta_{\FC} > 0$:
\begin{equation}
    \sum_{k=0}^{K-1} h\,\norm{\hat{o}^{\FC} - \ostarof{\tau_k}(\atauk{\tau_k})}
    \;\geq\; \delta_{\FC}.
    \label{eq:fc-persist}
\end{equation}
\end{propA}

\begin{proof}
Each term satisfies
$\norm{\hat{o}^{\FC} - \ostarof{\tau_k}(\atauk{\tau_k})} \geq \delta_{\FC}$
(by definition of infimum), and $\sum_{k=0}^{K-1} h = 1$.
\end{proof}

\begin{remA}
Fixed conditioning never applies the dynamics-alignment map $\Fof{\tau}$
and therefore cannot benefit from the contraction mechanism.
The conditioning mismatch is determined entirely by the quality of $\hat{o}^{\FC}$
and persists regardless of $K$.
\end{remA}

%-----------------------------------------------------------------------
\subsection{End-to-End Bounds}
\label{app:e2e}
%-----------------------------------------------------------------------

\begin{propA}[Conditioning Mismatch Implies Action Deviation]
\label{prop:gronwall}
Consider two action trajectories starting from the same noise $\atauk{\tau_0}$:
one using conditioning $\{\ohattauk{\tau_k}\}$ and one using the dynamics-aligned
conditioning $\{\ostarof{\tau_k}\}$, producing terminal actions $a_K$ and
$a^{\refsym}_K$ respectively.
Under Assumptions~\ref{ass:cross-lip}--\ref{ass:bounded}
and~\ref{ass:contraction} (the latter ensures the fixed-point map
$a \mapsto \ostarof{\tau}(a)$ is well-defined with finite Lipschitz constant
$\bar{L}_{o^*}$, see proof below):
\begin{equation}
    \norm{a_K - a^{\refsym}_K}
    \;\leq\; \Gamma \sum_{k=0}^{K-1} h\,\norm{\ohattauk{\tau_k} - \ostarof{\tau_k}(\atauk{\tau_k})},
    \label{eq:gronwall}
\end{equation}
where $\Gamma = e^{L_{\pi,a} + L_\pi \bar{L}_{o^*}} \cdot L_\pi$.
\end{propA}

\begin{proof}
Define $e_k := \norm{\atauk{\tau_k} - a^{\refsym}_k}$.
By the Euler update:
\begin{equation}
    e_{k+1}
    \leq e_k + h\norm{\uhata{\tau_k}(\atauk{\tau_k}, \tau_k \mid \ohattauk{\tau_k})
    - \uhata{\tau_k}(a^{\refsym}_k, \tau_k \mid \ostarof{\tau_k}(a^{\refsym}_k))}.
    \label{eq:ek-update}
\end{equation}
Decomposing the velocity difference:
\begin{align}
    &\uhata{\tau_k}(\atauk{\tau_k}, \tau_k \mid \ohattauk{\tau_k})
    - \uhata{\tau_k}(a^{\refsym}_k, \tau_k \mid \ostarof{\tau_k}(a^{\refsym}_k))
    \nonumber \\
    &= \bigl[\uhata{\tau_k}(\atauk{\tau_k}, \tau_k \mid \ohattauk{\tau_k})
    - \uhata{\tau_k}(a^{\refsym}_k, \tau_k \mid \ohattauk{\tau_k})\bigr]
    \nonumber \\
    &\quad + \bigl[\uhata{\tau_k}(a^{\refsym}_k, \tau_k \mid \ohattauk{\tau_k})
    - \uhata{\tau_k}(a^{\refsym}_k, \tau_k \mid \ostarof{\tau_k}(a^{\refsym}_k))\bigr].
    \label{eq:vel-decomp}
\end{align}
By Assumption~\ref{ass:state-lip}:
\begin{equation}
    \norm{\uhata{\tau_k}(\atauk{\tau_k}, \tau_k \mid \ohattauk{\tau_k})
    - \uhata{\tau_k}(a^{\refsym}_k, \tau_k \mid \ohattauk{\tau_k})}
    \leq L_{\pi,a}\,e_k.
    \label{eq:state-bound}
\end{equation}
By Assumption~\ref{ass:cross-lip} and the triangle inequality:
\begin{align}
    &\norm{\uhata{\tau_k}(a^{\refsym}_k, \tau_k \mid \ohattauk{\tau_k})
    - \uhata{\tau_k}(a^{\refsym}_k, \tau_k \mid \ostarof{\tau_k}(a^{\refsym}_k))}
    \nonumber \\
    &\quad\leq L_\pi\norm{\ohattauk{\tau_k} - \ostarof{\tau_k}(a^{\refsym}_k)}
    \leq L_\pi\norm{\ohattauk{\tau_k} - \ostarof{\tau_k}(\atauk{\tau_k})}
    + L_\pi\bar{L}_{o^*}\,e_k,
    \label{eq:cond-bound}
\end{align}
where $\bar{L}_{o^*}$ is the Lipschitz constant of
$a \mapsto \ostarof{\tau}(a)$, finite by the implicit function theorem applied
to $\Fof{\tau}(\hat{o}; a) = \hat{o}$ at its fixed point (whose well-posedness
requires the Jacobian $I - \nabla_{\hat{o}} \Fof{\tau}$ to be invertible---ensured
by Assumption~\ref{ass:contraction} via $\kappa^{(\tau)} < 1$).
Combining~\eqref{eq:state-bound} and~\eqref{eq:cond-bound}:
\begin{equation}
    e_{k+1} \leq
    \bigl(1 + h(L_{\pi,a} + L_\pi\bar{L}_{o^*})\bigr)\,e_k
    + hL_\pi\norm{\ohattauk{\tau_k} - \ostarof{\tau_k}(\atauk{\tau_k})}.
    \label{eq:ek-recurrence}
\end{equation}
The discrete Gronwall lemma with $e_0 = 0$ and
$(1+\lambda/K)^K \leq e^\lambda$ gives~\eqref{eq:gronwall}.
\end{proof}

\begin{corA}[End-to-End Bounds]
\label{cor:e2e}
\textbf{DAP (action error).}\;
Applying Proposition~\ref{prop:gronwall} with the DAP conditioning sequence
$\{\ohattauk{\tau_k}\}$ and Proposition~\ref{prop:dap-convergence}:
\begin{equation}
    \norm{\hat{a}^{\DAP} - a^{\refsym}_K}
    \leq \Gamma\,\eta_K.
    \label{eq:dap-to-ref}
\end{equation}
To compare the reference trajectory (learned model, dynamics-aligned conditioning)
against the expert, we apply Gronwall a second time.
The velocity difference between the learned reference and the ideal expert
trajectory decomposes into three sources:
\begin{align}
    &\uhata{\tau_k}(a^{\refsym}_k, \tau_k \mid \ostarof{\tau_k})
    - \uideala{\tau_k}(\atauk{\tau_k}, \tau_k \mid o^{\expt})
    \nonumber \\
    &= \underbrace{\bigl[\uhata{\tau_k}(a^{\refsym}_k, \tau_k \mid \ostarof{\tau_k})
    - \uhata{\tau_k}(\atauk{\tau_k}, \tau_k \mid \ostarof{\tau_k})\bigr]}_{\text{state difference: } \leq\, L_{\pi,a}\,\norm{a^{\refsym}_k - \atauk{\tau_k}}}
    \nonumber \\
    &\quad+ \underbrace{\bigl[\uhata{\tau_k}(\atauk{\tau_k}, \tau_k \mid \ostarof{\tau_k})
    - \uhata{\tau_k}(\atauk{\tau_k}, \tau_k \mid o^{\expt})\bigr]}_{\text{conditioning difference: } \leq\, L_\pi\,\Delta_{\tau_k}}
    \nonumber \\
    &\quad+ \underbrace{\bigl[\uhata{\tau_k}(\atauk{\tau_k}, \tau_k \mid o^{\expt})
    - \uideala{\tau_k}(\atauk{\tau_k}, \tau_k \mid o^{\expt})\bigr]}_{\text{direct model error: } \leq\, \varepsilon_\pi}.
    \label{eq:three-sources}
\end{align}
The first term uses Assumption~\ref{ass:state-lip}, the second uses
Assumption~\ref{ass:cross-lip} together with the bridge bound
$\norm{\ostarof{\tau_k} - o^{\expt}} \leq \Delta_{\tau_k}$
(Proposition~\ref{prop:bridge}), and the third uses
Assumption~\ref{ass:approx}.
Gronwall then gives, with $\Gamma' := e^{L_{\pi,a}}$,
\begin{equation}
    \norm{a^{\refsym}_K - a^{\expt}}
    \leq \Gamma' \Bigl(
    \varepsilon_\pi + L_\pi \sum_{k=0}^{K-1} h\,\Delta_{\tau_k}
    \Bigr)
    =: \Gamma'\,\mathcal{M}.
    \label{eq:ref-to-expert}
\end{equation}
Using $\Delta_{\tau_k} \leq (\varepsilon_f + L_f\varepsilon_\pi)
(1-\tau_k)/(1-L_\pi L_f)$
and $\sum_{k=0}^{K-1}h(1-\tau_k) = (K+1)/(2K)$:
\begin{equation}
    \mathcal{M} = \varepsilon_\pi
    + \frac{L_\pi(\varepsilon_f + L_f\,\varepsilon_\pi)}{1 - L_\pi L_f}
    \cdot \frac{K+1}{2K}.
    \label{eq:M-explicit}
\end{equation}
Note that $\varepsilon_\pi$ appears both directly (policy model error)
and indirectly through $\Delta_{\tau_k}$ (conditioning error induced by
imperfect policy).
By the triangle inequality:
\begin{equation}
    \norm{\hat{a}^{\DAP} - a^{\expt}}
    \leq \Gamma\,\eta_K + \Gamma'\,\mathcal{M}
    \;\xrightarrow{K\to\infty}\;
    \Gamma'\,\mathcal{M}.
    \label{eq:dap-full}
\end{equation}

\medskip\noindent
\textbf{DAP (future-observation error).}\;
The terminal future-observation prediction $\hat{o}^{\DAP}$ is the
extrapolation at the last DAP step $\tau_{K-1} = 1 - 1/K \in [0,1)$, namely
$\ohattauk{\tau_K} = \Fof{\tau_{K-1}}(\ohattauk{\tau_{K-1}})$.
By the triangle inequality, and since $\ostarof{\tau_{K-1}}$ is a fixed point of $\Fof{\tau_{K-1}}$:
\begin{equation}
    \norm{\hat{o}^{\DAP} - o^{\expt}}
    \;\leq\; \underbrace{\norm{\ohattauk{\tau_K} - \ostarof{\tau_{K-1}}}}_{\leq\,\kappa^{(\tau_{K-1})}\delta_{K-1}\,\leq\,\delta_{K-1}}
    \;+\; \underbrace{\norm{\ostarof{\tau_{K-1}} - o^{\expt}}}_{\leq\,\Delta_{\tau_{K-1}}}.
    \label{eq:dap-future}
\end{equation}
From Proposition~\ref{prop:dap-convergence}(b),
$\delta_{K-1} \leq (L_\pi L_f)^{K-1}\delta_0 + C_{\mathrm{drift}}\,h/(1-L_\pi L_f) = \mathcal{O}(1/K)$
(the exponential transient is dominated by the $\mathcal{O}(h)$ term).
From Proposition~\ref{prop:bridge} with $1 - \tau_{K-1} = 1/K$,
$\Delta_{\tau_{K-1}} = \mathcal{O}(1/K)$.
Therefore:
\begin{equation}
    \norm{\hat{o}^{\DAP} - o^{\expt}} \;\leq\; \mathcal{O}(1/K),
    \label{eq:dap-future-rate}
\end{equation}
matching the informal statement~\eqref{eq:main-dap}.

\medskip\noindent
\textbf{Fixed conditioning.}\;
The conditioning is $\ohattauk{\tau_k} = \hat{o}^{\FC}$ for all $k$.
By Proposition~\ref{prop:gronwall}:
\begin{equation}
    \norm{\hat{a}^{\FC} - a^{\refsym}_K}
    \leq \Gamma \sum_{k=0}^{K-1} h\norm{\hat{o}^{\FC} - \ostarof{\tau_k}(\atauk{\tau_k})}.
    \label{eq:fc-to-ref}
\end{equation}
By Proposition~\ref{prop:fc-persist}, the tracking sum is uniformly
lower-bounded by $\delta_{\FC}$ and therefore the right-hand side does not vanish
as $K \to \infty$.
Therefore:
\begin{equation}
    \norm{\hat{a}^{\FC} - a^{\expt}}
    \leq \Gamma \sum_{k=0}^{K-1} h\norm{\hat{o}^{\FC} - \ostarof{\tau_k}(\atauk{\tau_k})}
    + \Gamma'\,\mathcal{M},
    \label{eq:fc-full}
\end{equation}
where the tracking term (first summand) is uniformly lower-bounded by
$\Gamma\,\delta_{\FC} > 0$ and does not vanish with $K$.
Likewise, the future-observation surrogate $\hat{o}^{\FC}$ is held fixed and
does not approach $o^{\expt}$ as $K$ grows, so
$\norm{\hat{o}^{\FC} - o^{\expt}}$ is a positive constant independent of $K$.

\medskip\noindent
\textbf{Comparison.}\;
Both methods share the model error $\Gamma'\,\mathcal{M}$.
As $K \to \infty$:
\begin{itemize}[nosep]
    \item DAP: the tracking term $\Gamma\,\eta_K \to 0$.
    The action bound converges to pure model error, and the future-observation
    error vanishes at rate $1/K$.
    \item Fixed conditioning: the action-tracking term is uniformly
    lower-bounded by $\Gamma\,\delta_{\FC} > 0$, and the future-observation
    error is a positive constant independent of $K$.
\end{itemize}
DAP therefore admits a strictly tighter provable guarantee for
sufficiently large $K$.
\end{corA}

%-----------------------------------------------------------------------
\subsection{Measured Contraction Constants}
\label{app:measured}
%-----------------------------------------------------------------------

We estimate the cross-conditioning constants from the velocity fields used at inference.
At each flow time $\tau$, $L_\pi(\tau)$ is the largest ratio
\[
    \frac{\norm{\uhata{\tau}(a^{(\tau)}, \tau \mid \hat{o} + r\norm{\hat{o}}e) - \uhata{\tau}(a^{(\tau)}, \tau \mid \hat{o})}}{r\norm{\hat{o}}}
\]
over random unit directions $e$ and relative steps $r \in \{10^{-3}, 10^{-2}, 10^{-1}\}$, where $\hat{o}$ is DAP's current future-observation estimate, and $L_f(\tau)$ is defined analogously with the action estimate $\hat{a}$.
Relative steps keep the perturbation comparable between the VAE latent and normalized action spaces.
We use 200 held-out states per task, $K=10$ flow steps and 8 random directions, and average over states.
$L_\pi(\tau)L_f(\tau)$ stays below one at every $\tau$ on every task, so Assumption~\ref{ass:contraction} holds empirically for our models.
The resulting $\kappa^{(\tau)}$ is largest at $\tau=0$, with $0.055$ on PushCube-v1, $0.151$ on DrawTriangle-v1, $0.307$ on PickCube-v1 and $0.419$ on StackCube-v1, and falls to $0.001$--$0.002$ by $\tau=0.9$.
Figure~\ref{fig:alignment_error} shows DAP's alignment error during sampling.

\end{document}